\documentclass[11pt]{article}

\usepackage[final]{acl}

\usepackage{times}
\usepackage{latexsym}
\usepackage{amsmath}
\usepackage[T1]{fontenc}

\usepackage[utf8]{inputenc}
\usepackage{booktabs}
\usepackage{graphicx}
\usepackage{subcaption}
\usepackage{microtype}
\usepackage{amsfonts}
\usepackage{inconsolata}
\usepackage{multirow}

\usepackage{graphicx}
\usepackage{microtype}
\usepackage{amsmath}
\usepackage{extarrows}
\usepackage[most]{tcolorbox}
\usepackage{subcaption}
\usepackage{booktabs}
\usepackage[table]{xcolor}
\usepackage{amssymb}
\usepackage{pifont}
\usepackage[dvipsnames]{xcolor}
\newcommand{\upm}{\ensuremath{^{\scriptscriptstyle\uparrow}}}
\newcommand{\downm}{\ensuremath{^{\scriptscriptstyle\downarrow}}}

\newcommand{\HStar}[1]{\textcolor{#1}{\ding{73}}} 
\newcommand{\STri}[1]{\textcolor{#1}{\ensuremath{\blacktriangle}}} 
\newcommand{\SDot}[1]{\textcolor{#1}{\ensuremath{\bullet}}} 

\definecolor{myblue}{HTML}{104680}
\definecolor{myred}{HTML}{D32F2F}

\newcommand{\mstd}[2]{\ensuremath{#1\,{\scriptscriptstyle #2}}}

\title{Diagnosing Retrieval Bias Under Multiple In-Context Knowledge Updates in Large Language Models}

\author{
 \textbf{Boyu Qiao\textsuperscript{1,2}},
 \textbf{Sean Guo\textsuperscript{3}},
 \textbf{Xian Yang\textsuperscript{4}},
 \textbf{Kun Li\textsuperscript{1}},
\\
 \textbf{Wei Zhou\textsuperscript{1}},
 \textbf{Songlin Hu\textsuperscript{1,2}},
 \textbf{Yunya Song\textsuperscript{3}},
\\
\\
 \textsuperscript{1}Institute of Information Engineering, Chinese Academy of Sciences,
 \\
 \textsuperscript{2}School of Cyber Security, University of Chinese Academy of Sciences,
 \\
 \textsuperscript{3}Hong Kong University of Science and Technology,
 \textsuperscript{4}The University of Manchester,
\\
 \small{
   \textbf{Correspondence:} \href{yunyasong@ust.hk}{yunyasong@ust.hk}
 }
}

\begin{document}
\maketitle
\begin{abstract}


LLMs are widely used in knowledge-intensive tasks where the same fact may be revised multiple times within context. Unlike prior work focusing on one-shot updates or single conflicts, multi-update scenarios contain multiple historically valid versions that compete at retrieval, yet remain underexplored. This challenge resembles the AB-AC interference paradigm in cognitive psychology: when the same cue A is successively associated with B and C, the old and new associations compete during retrieval, leading to bias. Inspired by this, we introduce a Dynamic Knowledge Instance (DKI) evaluation framework, modeling multi-updates of the same fact as a cue paired with a sequence of updated values, and assess models via endpoint probing of the earliest (initial) and latest (current) states. Across diverse LLMs, we observe that retrieval bias intensifies as updates increase, earliest-state accuracy stays high while latest-state accuracy drops substantially. Diagnostic analyses of attention, hidden-state similarity, and output logits further reveal that these signals become flatter and weakly discriminative on errors, providing little stable basis for identifying the latest update. Finally, cognitively inspired heuristic intervention strategies  yield only modest gains and do not eliminate the bias. Our results reveal a persistent challenge in tracking and following knowledge updates in long contexts. Code at: \href{https://anonymous.4open.science/r/RetrivalBiasDKI-7CE6}{Repo}.

\end{abstract}

\section{Introduction}

Large language models (LLMs) are widely used in knowledge-intensive applications such as search assistants and knowledge-base question answering \cite{ma2025large}. In these scenarios, world knowledge continually evolves \cite{zhang2025self}, for example, changes in national leaders, corporate executives, or regulatory provisions. Ideally, deployed LLMs should stay aligned with up-to-date information to answer queries about the current state of a given fact, while still retaining the ability to recover its historical states when explicitly requested \cite{piryani2025s, zhu2025evolvebench}.

Prior work has studied LLM behavior under knowledge conflicts and updates from several angles \cite{xu2024knowledge}. One line constructs large-scale editing or conflict datasets \cite{su2024texttt, marjanovic2024dynamicqa} to analyze how models trade off parametric memory against contextual or retrieved evidence, and how stable this behavior is across tasks and editing operations \cite{tao2024context,sun2025taskmattersknowledgerequirements}. Another line develops temporally structured benchmarks that track evolving world knowledge and evaluate the impact of outdated parametric memory on performance \cite{meem2024pat,lin2025temporal}. However, most existing evaluations map each query to a single reference version of a fact, primarily capturing the single-update setting. In real-world deployments, the same fact may be revised multiple times within context, causing multiple versions to coexist and compete during retrieval. For this multi-version competition induced by multi-updates to the same cue, existing work still lacks a controlled evaluation framework, and it remains underexplored whether models exhibit retrieval bias between the current state and earlier historical states.

From a cross-disciplinary perspective, this multi-candidate competition echoes a classic phenomenon in cognitive psychology: AB-AC interference. In the AB-AC paradigm, participants first learn an association \(A\!:\!B\) and later learn a conflicting updated association \(A\!:\!C\) \citep{lewis2014competition,chanales2019interference}. During the testing phase, the old and new associations compete with each other in the retrieval process, thereby inducing retrieval bias. We adopt the AB-AC paradigm as a heuristic analytical lens to extract the essential structure of multiple in-context updates under a shared cue. Guided by this lens, we design controlled evaluations to probe retrieval outcomes under multi-candidate competition.



Concretely, we propose a Dynamic Knowledge Instance (DKI) evaluation framework, where each instance consists of a single semantic cue (e.g., President of Italy) and a sequence of updated values, forming a cue-value trajectory that effectively abstracts real-world systems such as databases and knowledge graphs. Inspired by the AB-AC interference paradigm, we adopt endpoint probing as a minimal yet informative probe: by querying the earliest (initial) and latest (current) states, we test whether models preserve historical access while tracking the most recent update. Experiments on a broad set of LLMs reveal a retrieval bias: earliest-state accuracy stays high while latest-state accuracy drops substantially as updates increase, yielding a widening earliest-latest accuracy gap. To go beyond black-box measurements, we further analyze the model's internal signals, including attention allocation, hidden-state similarity, and output logits, to reveal how these signals align with each candidate update. Finally, inspired by heuristic memory strategies in cognitive psychology, we develop prompt-based intervention strategies as directional attempts to mitigate this bias. They achieve limited progress in latest-state retrieval yet cannot eliminate the fundamental retrieval bias. This outcome highlights the core challenge of reliable in-context knowledge tracking in long-context, multi-version competitive settings. Our contributions are as follows:

\noindent \textbf{Framework.} Inspired by the AB-AC interference paradigm, we propose the DKI evaluation framework that formalizes multi-updates of the same fact, and we adopt endpoint probing to controllably evaluate the earliest and latest states, thereby quantifying LLMs' retrieval bias in multi-update scenarios.



\noindent \textbf{Findings.} Across diverse LLMs, we observe an earliest-latest accuracy gap that widens as updates increase. Analyses of attention, hidden-state similarity, and logits show that the internal signals are sharply defined in correct cases but become flattened in error cases, offering weak discriminative evidence for identifying the newest update.


\noindent \textbf{Interventions.} We translate several cognitively motivated heuristics into prompt-based interventions. These interventions yield only modest gains in latest-state retrieval and do not eliminate the bias, thus highlighting the limits of generic prompting and motivating the development of more targeted update-tracking mechanisms.

\section{Related Works}

\vspace{-5pt}
\subsection{LLMs Knowledge Updates and Conflicts}

Existing work has examined how LLMs behave under knowledge updating and conflicts from multiple angles \cite{xu2024knowledge}. Some studies build large-scale editing or conflict datasets \cite{marjanovic2024dynamicqa} to investigate how models trade off parametric memory against contextual or retrieved evidence. For instance, ConflictBank \cite{su2024texttt} systematically characterizes model outputs when parametric knowledge conflicts with retrieved evidence. Jin et al. \shortcite{jin2024tug} and Sun et al. \shortcite{sun2025taskmattersknowledgerequirements} analyze how reliance on parameters versus external documents varies with context and task, and propose resolving conflicts via contrastive decoding or local knowledge editing. Other work introduces temporally structured benchmarks to track how evolving world knowledge and outdated parametric memory affect performance \cite{lin2025temporal}. For example, PAT \cite{meem2024pat} evaluates whether a model can return the up-to-date answer for a given time point after factual changes. However, most evaluations assume one update per query and rarely consider multi-updates. In this setting, models must choose among competing knowledge versions. Beyond reporting retrieval accuracy, we additionally analyze how internal signals vary across different updates to gain further insight into retrieval bias.


\vspace{-5pt}
\subsection{Cognitive Foundations: AB - AC Interference Paradigm}

In cognitive psychology, AB-AC paradigms study memory competition when the same cue is bound to conflicting associations (A-B then A-C), often yielding interference and systematic retrieval bias \cite{wahlheim2015testing,chanales2019interference}. We view our setting as a natural generalization from a single conflicting update to a sequence of repeated updates, where multiple candidates may compete at retrieval time. Cue-overload theory\cite{watkins1975buildup, burton2017associative} explains these effects: as more traces are linked to the same retrieval cue, competition increases and the probability of retrieving any single trace decreases. Moreover, subsequent work shows that access to old and new associations is shaped by encoding and retrieval strategies rather than simple overwriting. For example, rehearsal and elaborative semantic integration can strengthen target associations and improve recall \cite{cepeda2006distributed,craik1975depth, wahlheim2015testing}.




\begin{figure*}[htb]
\centering
\includegraphics[width=0.87\textwidth]{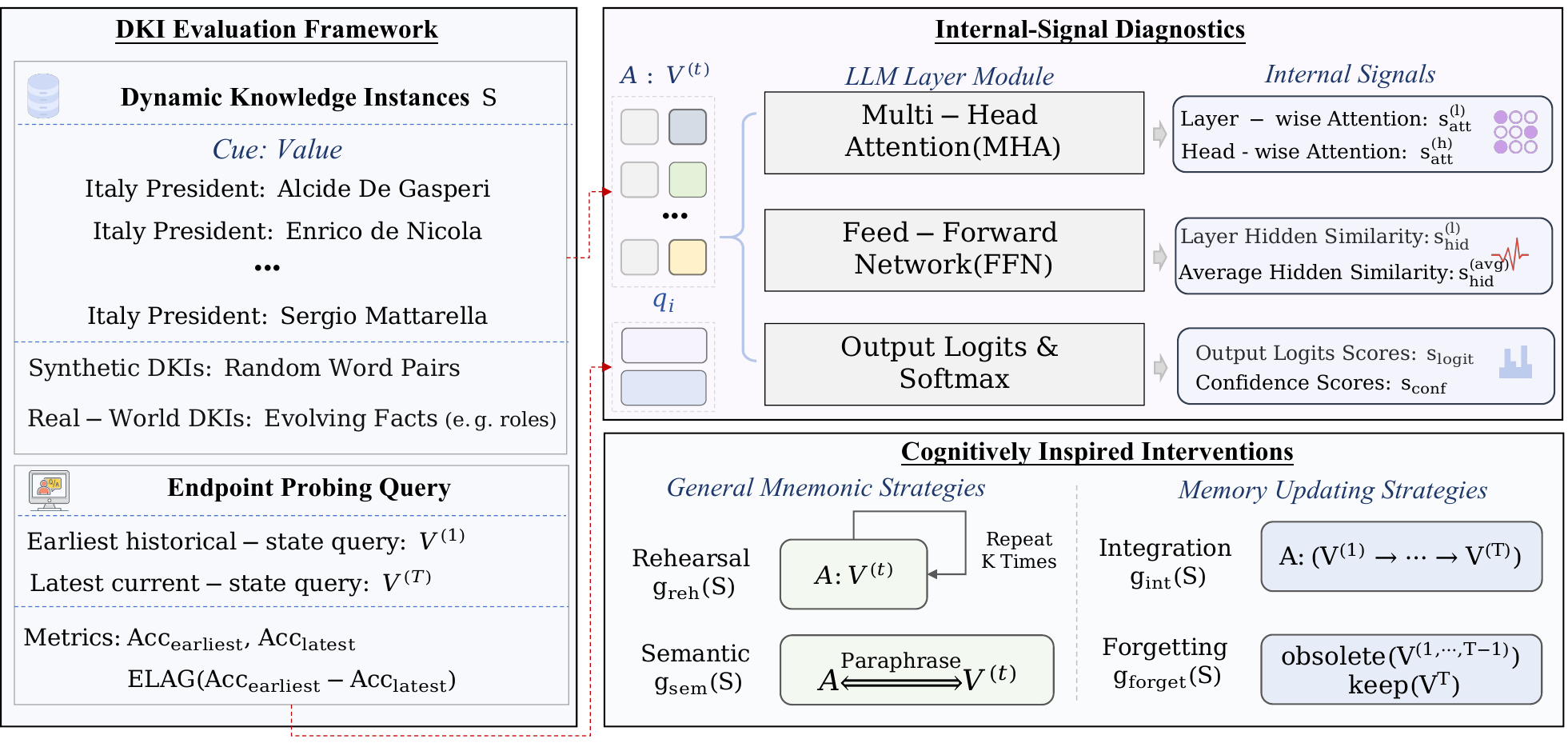}
\centering
\caption{Overview of the DKI evaluation framework, internal-signal diagnostics, and cognitively inspired intervention strategies for multi-update knowledge retrieval.}
\label{fig:1}
\end{figure*}

\section{Methodology}

To systematically investigate retrieval behavior under multi-updates of the same cue in long contexts, we propose a controlled evaluation framework, perform internal-signal diagnostics, and develop intervention strategies. An overview of our study design is shown in Figure~\ref{fig:1}. First, inspired by the AB-AC paradigm in cognitive psychology, we propose the Dynamic Knowledge Instance (DKI) evaluation framework. Building on this framework and drawing on the cue-overload theory \cite{burton2017associative}, we characterize the systematic retrieval bias exhibited by LLMs under multi-updates to the same cue, and how this bias scales with the number of updates. Second, by analyzing the alignment patterns and separability of attention allocation, hidden-state similarity, and output logits across candidate updates, we investigate whether these internal signals provide stable and discriminative local evidence that reliably points to the latest update. Finally, we translate general mnemonic and memory-updating strategies from cognitive theory into interventions, and evaluate their effectiveness and limitations in mitigating retrieval bias.





\subsection{Dynamic Knowledge Instance Evaluation Framework}

\subsubsection{Problem Setup and Notation}

\textbf{AB-AC interference paradigm.} The classical AB-AC paradigm in cognitive psychology examines retrieval interference and bias when the same cue is paired with competing associates \cite{lewis2014competition}.  In the classic experiment, participants first learn an association $A\!:\!B$ (AB phase) and then learn an updated, conflicting association $A\!:\!C$ (AC phase). 
In the test phase, both the original (AB) and the updated (AC) associations can be probed to assess recall of old versus new association.

\noindent \textbf{Real-world fact updates.} Real-world facts, however, often undergo more than a single update. Drawing on the cue-overload theory, that binding more traces to the same cue intensifies competition and makes retrieval increasingly difficult, we generalize the single-update AB-AC setting to multiple updates for the same cue and systematically probe how retrieval behaves as the update length grows. 
For a cue $A$, we define a dynamic knowledge instance (DKI) $S$ as a cue-value trajectory of length $T$:
\vspace{-7pt}
\begin{equation}
  \label{eq:1}
  S = A: V^{(1)} \cdots \Rightarrow A: V^{(t)}  \cdots \Rightarrow A: V^{(T)},
\end{equation}
where $V^{(t)}$ denotes the value associated with $A$ at update round $t$.

\subsubsection{Endpoint Probing}
Inspired by the old-new associative retrieval mechanism of the AB-AC paradigm, we adopt endpoint probing as a minimal controlled contrast of retrieval under multi-updates. This choice mirrors the AB-AC old-vs-new contrast and avoids ambiguity for intermediate states, enabling a cleaner test of whether retrieval is biased toward historical or current updates.
Formally, given a DKI trajectory $S$, we probe the two endpoints:
\begin{itemize}
  \vspace{-3pt}
  \item \textbf{Earliest historical-state query:} ask the model to output the initial value $V^{(1)}$.
  \vspace{-3pt}
  \item \textbf{Latest current-state query:} ask the model to output the most recent value $V^{(T)}$.
\end{itemize}
\vspace{-3pt}
By varying $T$, we quantify how retrieval of the earliest versus the latest state changes as updates increase, and we use the earliest-latest accuracy gap (ELAG) as an operational measure of retrieval bias under multiple in-context updates.

\subsubsection{DKIs Construction}
To instantiate the formal setting in controlled experiments, we construct DKIs from two complementary sources: synthetic knowledge associations inspired by classic AB-AC, and real-world knowledge associations.

\noindent  \textbf{Synthetic knowledge associations}. To precisely control the update length $T$ and reduce interference from parametric factual priors, we construct semantically arbitrary synthetic cue-value bindings. This design follows the arbitrary paired-associate paradigm commonly used in classic AB-AC experiments, which minimizes semantic priors and helps isolate the pure effects of interference and competition during retrieval. Concretely, we randomly sample the cue $A$ and values $\{V^{(t)}\}_{t=1}^{T}$ from English word forms \cite{balota2007english,ebert2009proactive}. 
Since these cue-value bindings do not correspond to stable real-world relations, answering primarily requires tracking the in-context update sequence rather than relying on memorized factual knowledge. This synthetic construction allows us to vary $T$ in a controlled manner and quantify how retrieval bias between early versus late states evolves under updates increase.

\noindent  \textbf{Real-world knowledge associations}. To examine whether a consistent retrieval bias also appears in realistic settings, we further evaluate on real-world multi-round fact updates. In this setting, the cue $A$ corresponds to a persistent semantic role or relation whose associated value changes over time. We adopt the evolving-knowledge dataset from EvolveBench \cite{zhu2025evolvebench}, which was originally proposed to assess LLMs' temporal awareness, and further clean and reorganize it. We extract its attribute-value sequence and reformat it into a DKI trajectory. We then apply the same endpoint probing protocol to real-world DKIs. 


Appendix \ref{Appendix:B} shows the prompt examples.

\subsection{Internal-Signal Diagnostics}

To connect the endpoint retrieval results of models with their internal computation processes, we examine attention allocation, hidden-state similarity, and output logits as internal diagnostic signals. We then use these indicators to quantify the model's relative preference among candidates when generating answers. Specifically, given a query $q_i$ and candidate values $\{V^{(t)}\}_{t=1}^{T}$, we extract activations at the answer position $p_{\mathrm{ans}}$, defined as the start token of the target value span in the output. For each layer $l$ and head $h$, we record the post-softmax attention weights $A^{(l,h)} \in \mathbb{R}^{M \times M}$ and hidden states $H^{(l)} \in \mathbb{R}^{M \times D}$, and the final-layer logits $z \in \mathbb{R}^{|\mathcal{V}|}$, where $M$ is the sequence length, $D$ is the hidden dimension, and $|\mathcal{V}|$ is the vocabulary size. Let $P_t$ denote the token set of candidate value $V^{(t)}$. Based on these activations, we derive three types of local scores for each $V^{(t)}$, and evaluate their alignment with the target correct update.

\noindent  \textbf{Attention-based scores}. For layer $l$ and attention head $h$, we extract the attention weights from the answer position $p_{\mathrm{ans}}$ to the token span $P_t$ of candidate values, and aggregate them via a length-normalized mean over the span to obtain the attention score for each candidate $V^{(t)}$:
\vspace{-6pt}
\begin{equation}
s_{\mathrm{att}}^{(l,h)}(t)=\frac{1}{|P_t|}\sum_{v\in P_t}A^{(l,h)}_{p_{\mathrm{ans}},v}.
\end{equation}
\noindent The layer-wise attention score is obtained by mean pooling over all heads within the same layer:
\vspace{-6pt}
\begin{equation}
s_{\mathrm{att}}^{(l)}(t)=\frac{1}{H}\sum_{h=1}^{H}s_{\mathrm{att}}^{(l,h)}(t),
\end{equation}
\noindent where $H$ is the number of heads per layer. The head-wise attention score is obtained by mean pooling across layers for the same head index:
\vspace{-6pt}
\begin{equation}
s_{\mathrm{att}}^{(h)}(t)=\frac{1}{L}\sum_{l=1}^{L}s_{\mathrm{att}}^{(l,h)}(t),
\end{equation}
\noindent where $L$ is the number of layers. These scores quantify the relative attention allocated to different candidate values during answer generation, enabling us to identify layers or head indices whose attention is most aligned with the correct update.

\noindent  \textbf{Hidden-state similarity scores}. We further leverage hidden state similarity to quantify the alignment between the model's internal representations and each candidate value $V^{(t)}$. For each layer $l$, we perform mean pooling on the hidden states over $P_t$ of $V^{(t)}$, obtaining representation:
\vspace{-6pt}
\begin{equation}
e_{\mathrm{cand}}^{(l)}(t)=\frac{1}{|P_t|}\sum_{v\in P_t}H^{(l)}_{v,:}.
\end{equation}
\noindent  Let $h_{\mathrm{ans}}^{(l)} = H^{(l)}_{p_{\mathrm{ans}}, :}$ denote the hidden state at the answer position $p_{\mathrm{ans}}$. We define the hidden-state similarity score for $V^{(t)}$ using cosine similarity:
\vspace{-6pt}
\begin{equation}
s_{\mathrm{hid}}^{(l)}(t)=\operatorname{Cosine}(e_{\mathrm{cand}}^{(l)}(t),h_{\mathrm{ans}}^{(l)}).
\end{equation}
\noindent  To obtain a more stable alignment signal across layers, we  compute the multi-layer average similarity score via aggregation:
\vspace{-6pt}
\begin{equation}
\label{eq:7}
s_{\mathrm{hid}}^{(\mathrm{avg})}(t)
= \frac{1}{L}\sum_{l=1}^{L} s_{\mathrm{hid}}^{(l)}(t).
\end{equation}
\noindent  This aggregated score captures the model's overall preference among candidate values in its representational space and is used as a local metric for hidden-state alignment.

\noindent  \textbf{Logit-based scores}. To characterize the model's confidence during answer generation, we record the output-layer logits and compute confidence scores. For a candidate $V^{(t)}$ that spans multiple tokens $P_t$, we read the output-layer logits for each token and average these logits to obtain the logit-based score for each candidate value:
\vspace{-6pt}
\begin{equation}
s_{\mathrm{logit}}(t)=\frac{1}{|P_t|}\sum_{v \in P_t} z_{v},
\end{equation}
where $z_{v}$ denotes the logit of vocabulary token $v$. To obtain a probabilistic confidence measure, we apply a softmax to $z$ to get the distribution $p = softmax(z)$, and similarly average the probabilities $P_t$ to define the confidence score for $V^{(t)}$:
\vspace{-6pt}
\begin{equation}
  \label{eq:9}
s_{\mathrm{conf}}(t)=\frac{1}{|P_t|}\sum_{v\in P_t} p_{v}.
\end{equation}
The score $s_{\mathrm{conf}}(t)$ provides a confidence score, which we use to analyze whether the model exhibits overconfidence when answering.

\subsection{Cognitively Inspired Interventions}
To reduce retrieval bias under DKIs ($S$, Eq.~(\ref{eq:1})), we heuristically translate insights from cognitive psychology into prompt-based interventions that can be directly applied to and evaluated on LLMs. We consider two categories of interventions $g_{\pi}(S)$: (i) general mnemonic strategies, including rote rehearsal ($g_{\text{reh}}$) \cite{glenberg1978type} and semantic elaboration ($g_{\text{sem}}$) based on levels-of-processing theory \cite{craik1972levels}. These strategies mainly improve the accessibility of new values to the model by enhancing encoding strength; and (ii) memory-updating strategies including memory integration ($g_{\text{int}}$) \cite{chanales2019interference, schlichting2015memory, guo2025providing} and directed forgetting ($g_{\text{forget}}$) \cite{macleod1999item}, which target same-cue multi-version competition, mitigate outdated-information interference, and optimize latest-state tracking.

\noindent \textbf{General mnemonic strategies.} 
(i) \textbf{Rote rehearsal} instructs the model to internally rehearse each newly read cue-value pair $K$ times:
$g_{\text{reh}}(S)= \underbrace{\big[ A:V^{(1)}, A:V^{(2)}, \cdots, A:V^{(T)} \big]}_{K\ \text{times}}$.
(ii) \textbf{Semantic elaboration} prompts the model to generate a semantic reformulation for each update record,
$g_{\text{sem}}(S)=\phi(A{:}V^{(t)})$,
where $\phi(\cdot)$ asks the model to elaborate the pair at a semantic level.

\noindent \textbf{Memory-updating strategies.}
(i) \textbf{Memory integration} prompts models to treat the cue-value pairs as a updated chain instead of independent pairs. Specifically, it organizes all updates into an explicit update chain,
$g_{\text{int}}(S)=[A:(V^{(1)} \rightarrow V^{(2)} \cdots \rightarrow V^{(T)})]$, encouraging the model to co-activate earlier and later values, thereby reducing the likelihood that outdated values are predicted.
(ii) \textbf{Directed forgetting} down-weights historical candidate values,
$g_{\text{forget}}(S)=[obsolete(V^{(1)},\ldots,V^{(T-1)}), keep(V^{(T)})]$, prompting the model to label earlier values as obsolete while reinforcing the most recent value, making outdated candidates less likely to be retrieved when multiple candidates coexist.

Appendix~\ref{Appendix:D.1} shows the prompt templates.

\begin{figure*}[htbp]
\centering
\includegraphics[width=0.80\textwidth]{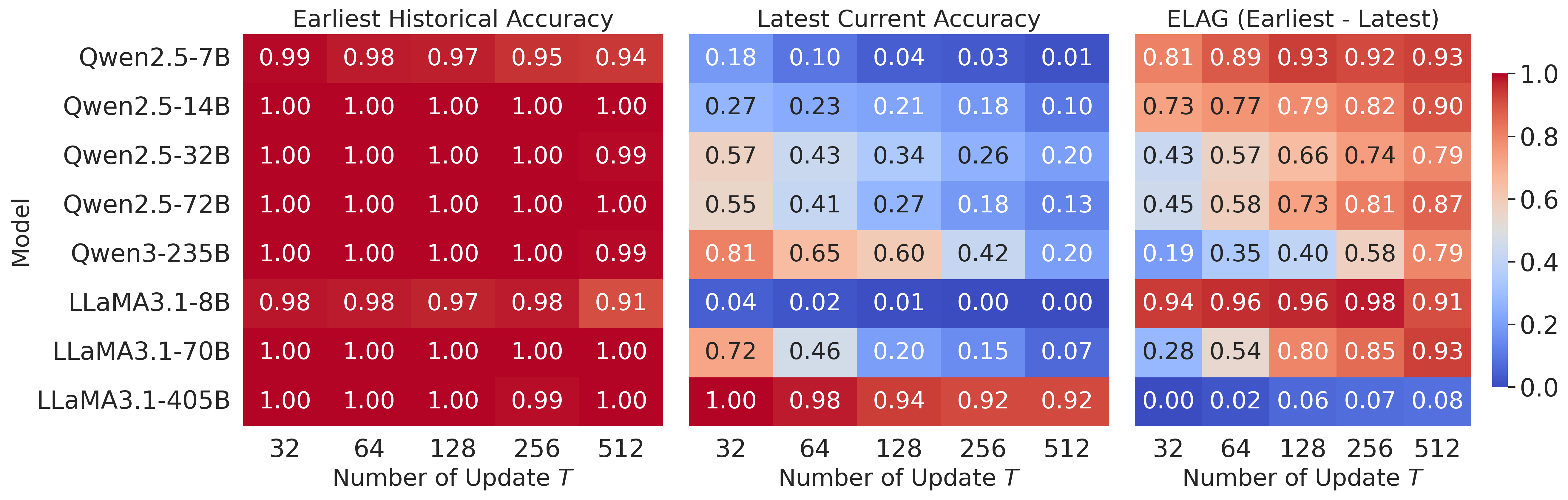}
\centering
\vspace{-0.2em}
\caption{Endpoint-probing performance on synthetic DKIs for LLaMA-3.1 and Qwen-2.5/3 model families. Left/middle: earliest and latest state accuracy. right: earliest-latest accuracy gap (ELAG) ($Acc_{earliest} - Acc_{latest}$).}
\label{fig:2}
\end{figure*}

\vspace{-5pt}
\section{Experiments}
\vspace{-5pt}
\subsection{Experiment Setup}

Table~\ref{tab:1} in Appendix~\ref{Appendix:A} summarizes our experimental setup. 
We evaluate the performance using both synthetic and real-world DKI datasets, covering widely used commercial and open-source LLMs. For brevity, we provide an abbreviated identifier for each model in parentheses and use these short forms throughout the figures and analysis. Our experimental analysis consists of five components: (\ref{sec:4.2}) Retrieval Bias Phenomenon Identification; (\ref{sec:4.3}) Real-World Data Validation; (\ref{sec:4.4}) Error Characteristic Analysis; (\ref{sec:4.5}) Internal Signal Diagnosis; (\ref{sec:4.6}) Heuristic Interventions.


\begin{figure}[htbp]
\centering
\includegraphics[width=0.45\textwidth]{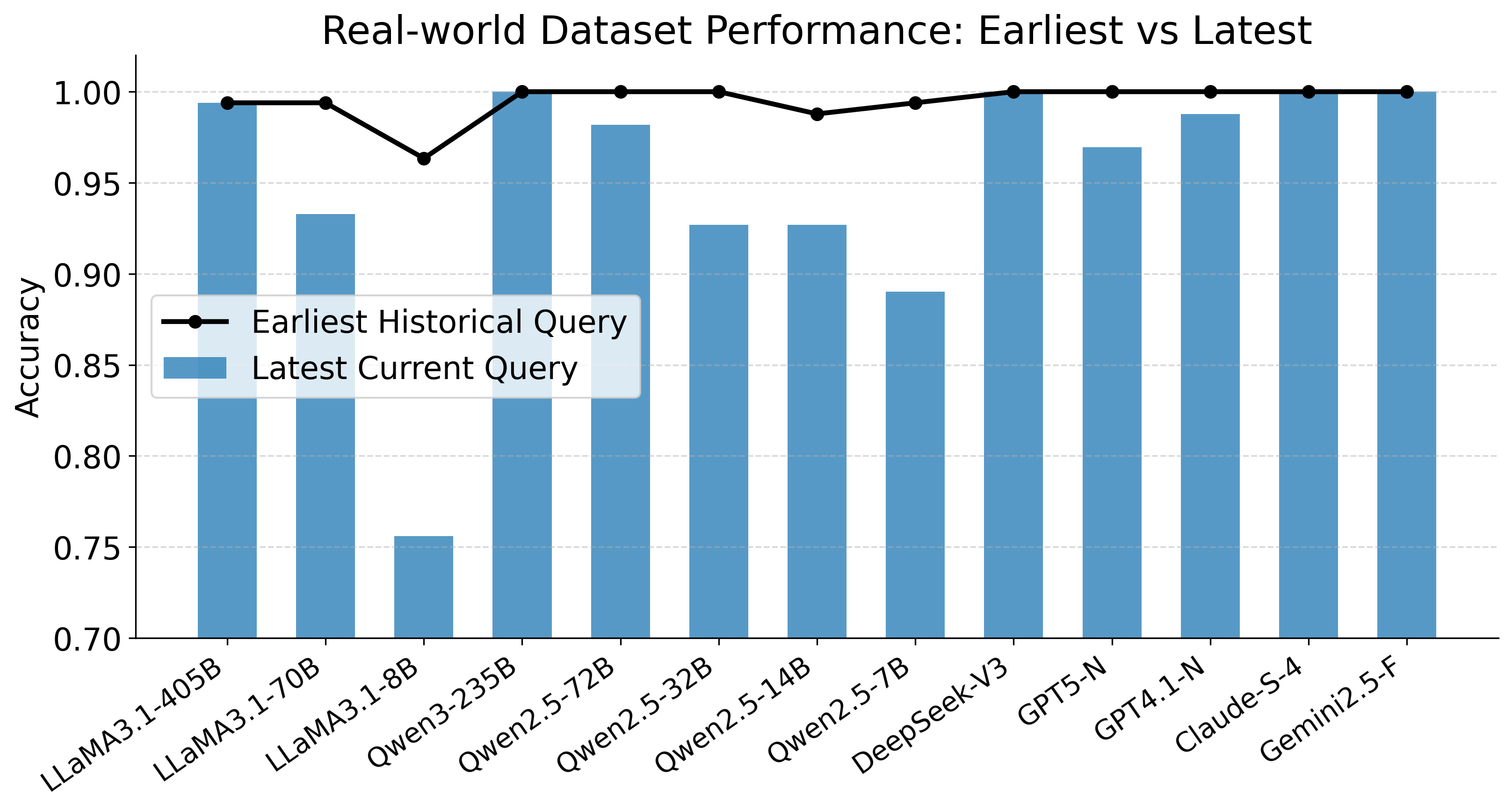}
\centering
\caption{Performance on real-world DKIs.}
\label{fig:5}
\end{figure}
\vspace{-5pt}
\subsection{Retrieval Bias Phenomenon Identification}
\label{sec:4.2}
Experiments across multiple LLMs show that the \textbf{synthetic DKI task exhibits a pronounced retrieval bias under in-context multi-candidate competition}. Figure~\ref{fig:2} summarizes the performance of the LLaMA-3.1 and Qwen-2.5/3 model familiesas as the update length $T$ increases from $32$ to $512$. Across models, earliest-state retrieval accuracy remains consistently high, whereas latest-state retrieval accuracy steadily degrades with longer update trajectories, yielding an increasingly large ELAG. Further analysis of model scale indicates that smaller models exhibit stronger retrieval bias, especially as the number of updates grows. Appendix~\ref{Appendix:C.1} (Figure~\ref{fig:3}) reports additional experiments on commercial LLMs, where the retrieval bias is also observed. The ELAG curve in Figure~\ref{fig:4} indicates that the gap expands rapidly as $T$ increases from $32$ to $128$, and then tends to saturate when $T \ge 256$. Overall, these results confirm that the retrieval bias is prevalent across different LLMs.


\vspace{-5pt}
\subsection{Real-World Data Validation}
\label{sec:4.3}
\textbf{The experimental results on real-world DKIs exhibit the same pattern observed in synthetic DKIs: retrieval bias persists.} Figure~\ref{fig:5} reports model accuracy on earliest-state and latest-state queries in the real-world DKI data. Although the average update length in this dataset is only $8.77$ (Table~\ref{tab:1}), far shorter than the update scales studied in the synthetic setting ($T=32\sim512$), we still observe a pronounced ELAG. This suggests that even under a small number of real-world updates, the model's ability to track the the latest state has already become unreliable. To better approximate realistic usage and move beyond the limitations of the structured cue:value format, we use GPT-4.1-N to rewrite real-world cue-value update trajectories into narrative long-text documents, embedding successive factual updates naturally within a coherent story context. Appendix~\ref{Appendix:C.2} provides concrete rewriting examples, and Figure~\ref{fig:6} reports results under this long-text format. While absolute performance differs from the structured setting, likely due to the combined effects of parametric priors and narrative distractions, the ELAG remains clearly observable, indicating that the retrieval bias is robust to changes in input format.


\vspace{-5pt}
\subsection{Error Characteristic Analysis}
\label{sec:4.4}
\textbf{Models exhibit markedly different failure modes on latest-state queries.} Figure~\ref{fig:7} presents the distribution of predicted answers over candidate updates for two representative open-source models, Qwen2.5-7B and LLaMA3.1-8B, with the update length at $T=32$. The two models display qualitatively distinct error patterns: LLaMA3.1-8B's wrong answers tend to drift and regress toward earlier candidate positions, whereas Qwen2.5-7B's predictions cluster near the sequence end but yield out-of-field (OOF) outputs. This behavioral characterization indicates that errors on latest-state queries are not homogeneous, and the underpinning failure mechanisms are strongly model-specific.



\begin{figure}[t]
    \centering
    \begin{subfigure}[t]{0.49\linewidth}
        \centering
        \includegraphics[width=\linewidth]{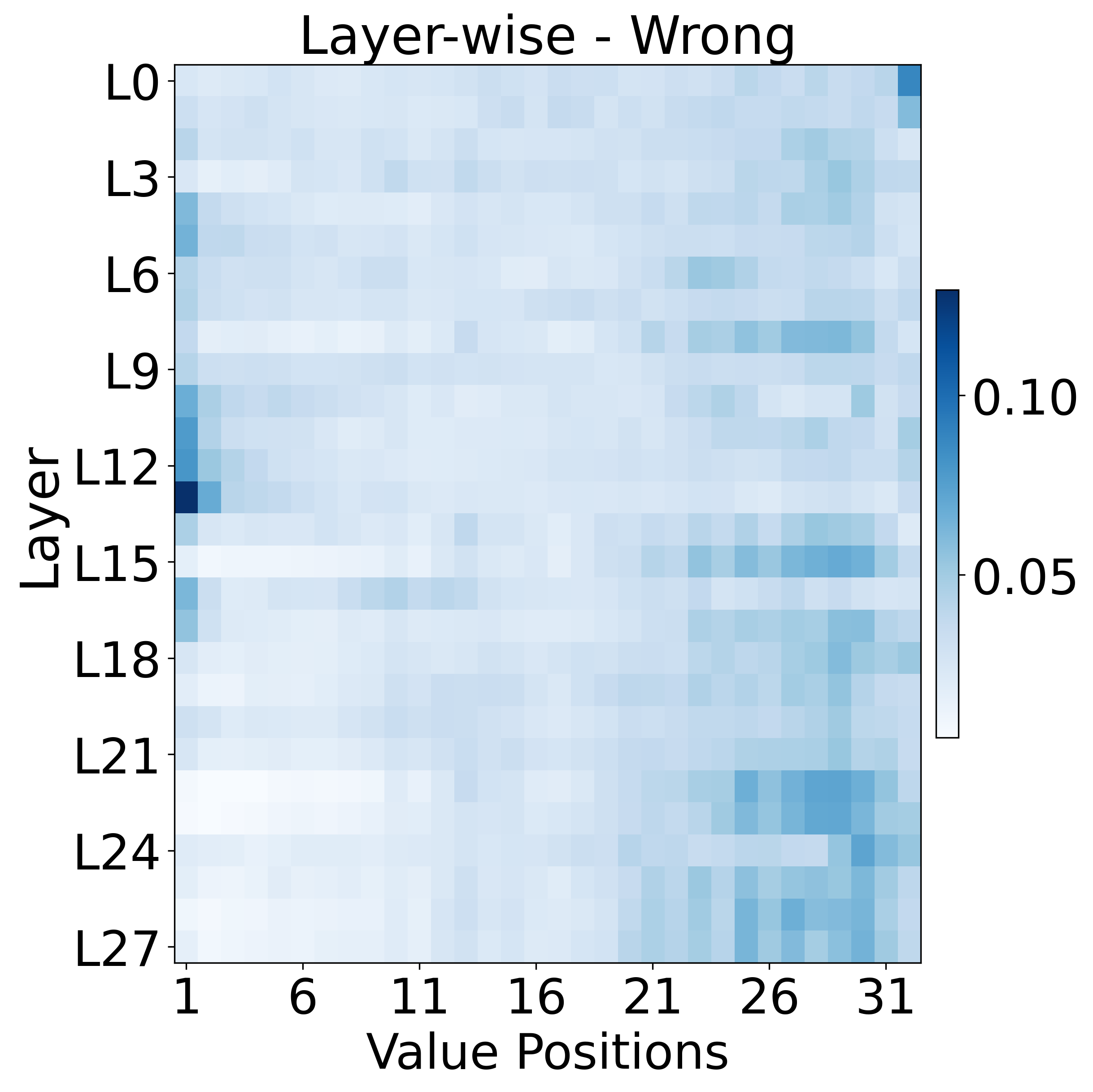}
        \caption{Qwen: Layer-wise.}
        \label{fig:att_wrong_qwen_layer}
    \end{subfigure}\hfill
    \begin{subfigure}[t]{0.49\linewidth}
        \centering
        \includegraphics[width=\linewidth]{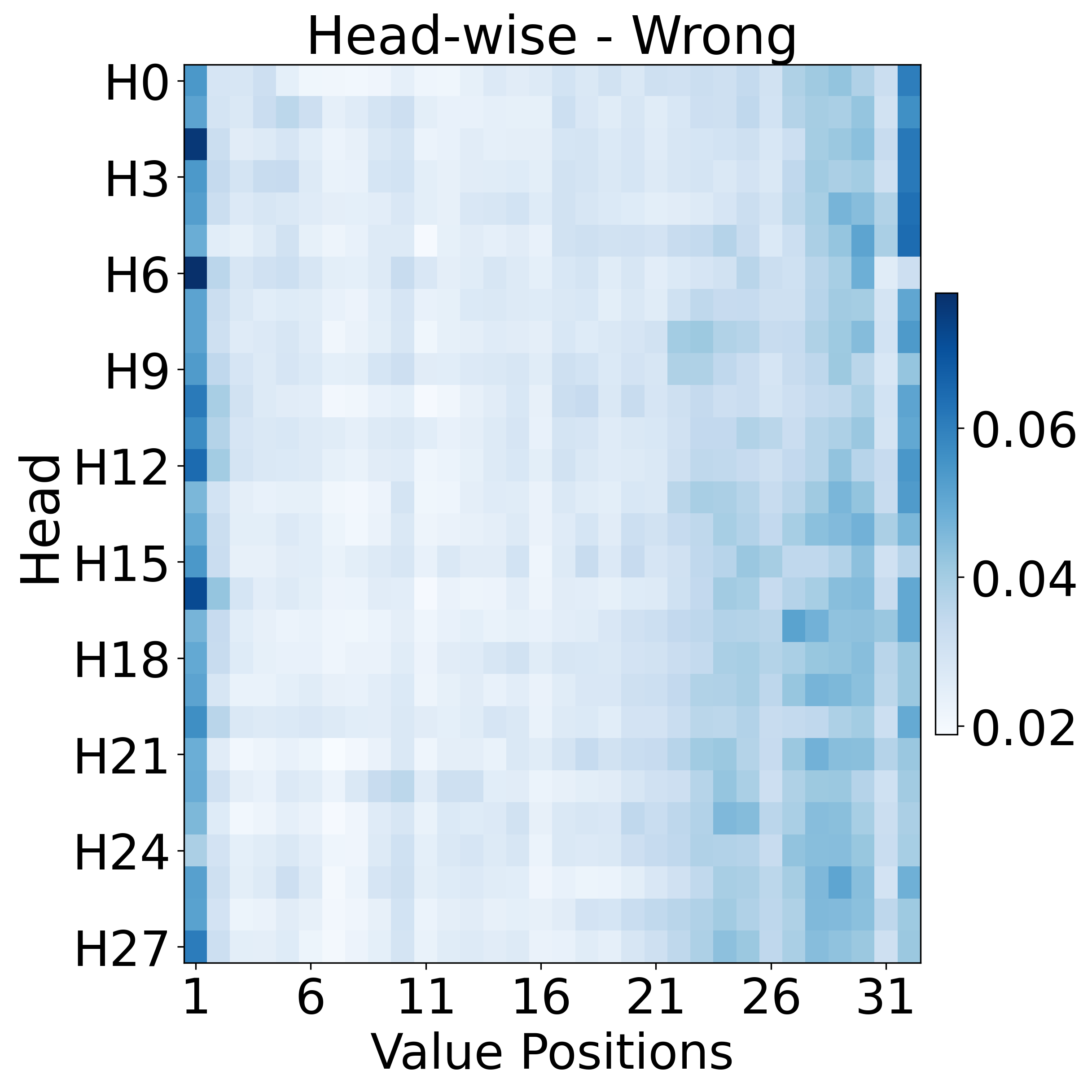}
        \caption{Qwen: Head-wise.}
        \label{fig:att_wrong_qwen_head}
    \end{subfigure}

    \vspace{-0.2em}

    \begin{subfigure}[t]{0.49\linewidth}
        \centering
        \includegraphics[width=\linewidth]{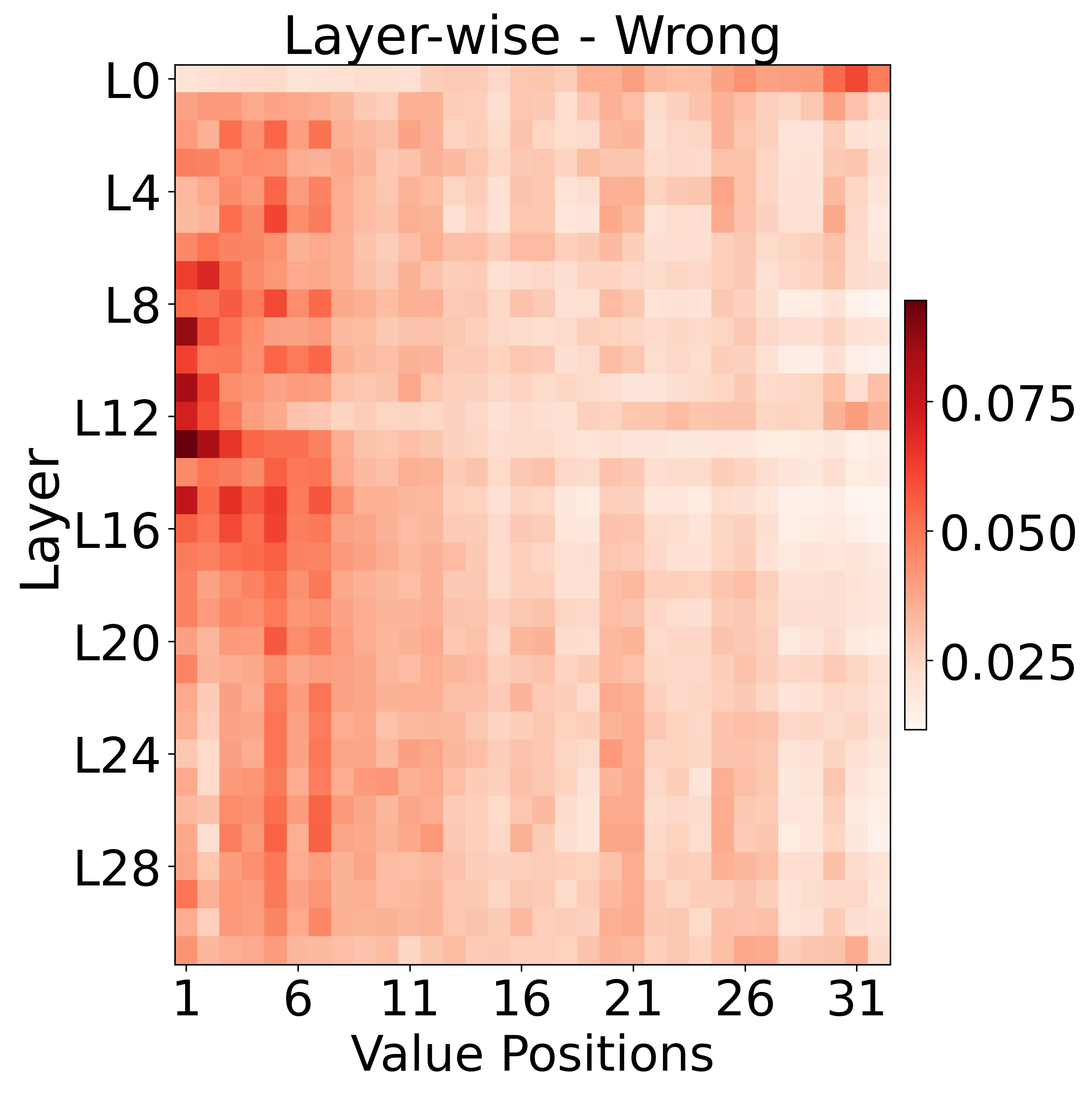}
        \caption{LLaMA: Layer-wise.}
        \label{fig:att_wrong_llama_layer}
    \end{subfigure}\hfill
    \begin{subfigure}[t]{0.49\linewidth}
        \centering
        \includegraphics[width=\linewidth]{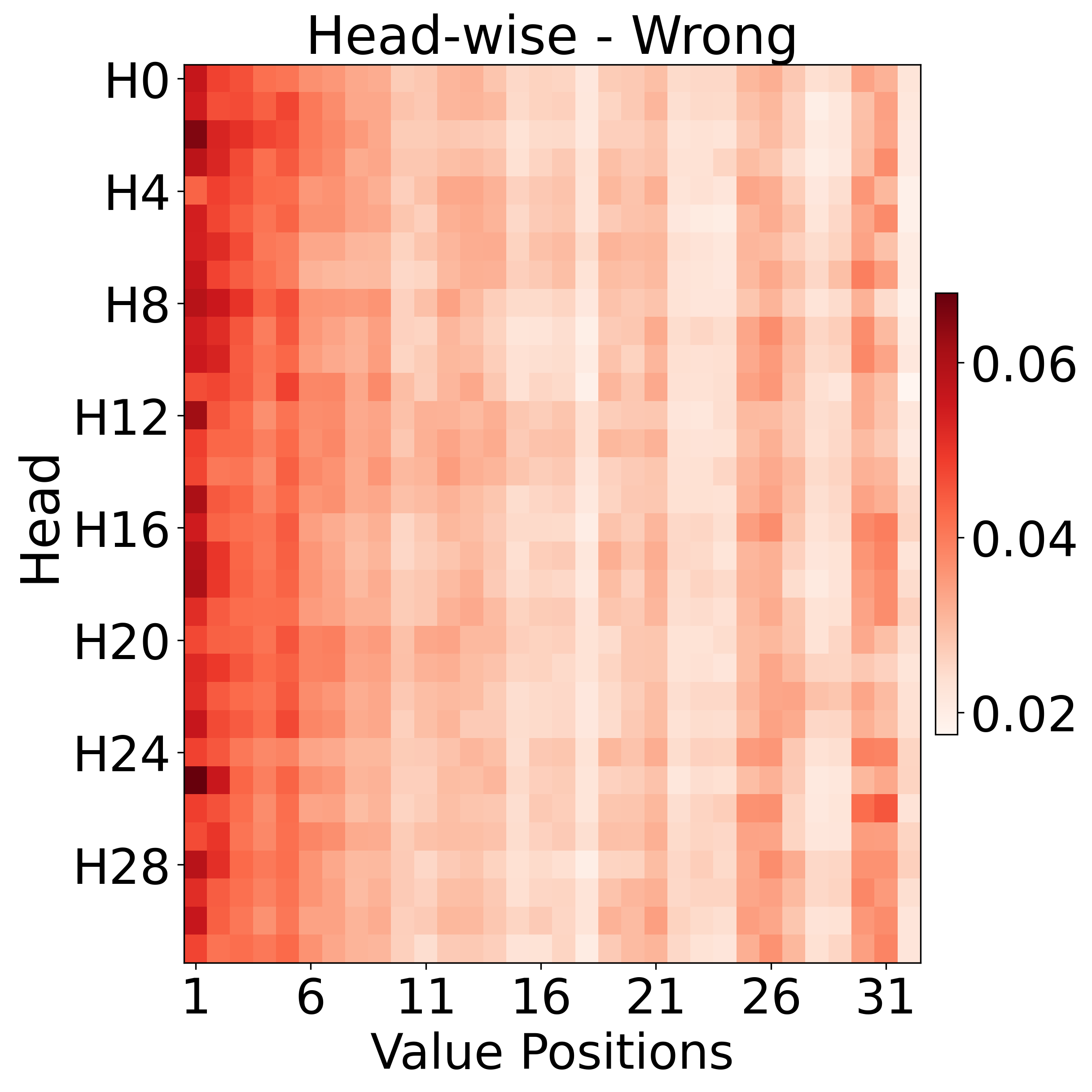}
        \caption{LLaMA: Head-wise.}
        \label{fig:att_wrong_llama_head}
    \end{subfigure}

    \vspace{-0.4em}
    \caption{Layer-wise and head-wise attention-score over candidate positions (x-axis) on wrong samples.}
    \label{fig:attn_wrong_all}
\end{figure}


\vspace{-5pt}
\subsection{Internal Signal Diagnosis}
\label{sec:4.5}

For Qwen2.5-7B and LLaMA3.1-8B at update length $T=32$, we split samples into correct/wrong groups based on whether they answer the latest-state query correctly. We then compare the attention, hidden-state similarity, and output logits between the two groups to analyze alignment of these three local signals with the correct answer.


\noindent \textbf{1. Attention-based scores show model-specific failure patterns on errors.} Figures~\ref{fig:attn_wrong_all} and \ref{fig:att_correct_all} present the layer-wise and head-wise attention-score heatmaps for the two models on wrong and correct samples, respectively. The attention distributions are similar across the two views and closely correspond to the predicted answer-position distributions (Figure~\ref{fig:7}), indicating attention scores partially reflect models' candidate-position preferences. Specifically, Qwen's attention mass concentrates near sequence tails but spreads to adjacent candidates, denoting coarse tail localization without precise latest-value alignment. In contrast, LLaMA places more attention to earlier candidates and maintains low attention on later updates, matching its weaker latest-state retrieval performance.

\begin{figure}[t]
    \centering
    \begin{subfigure}{0.85\linewidth}
        \centering
        \includegraphics[width=\linewidth]{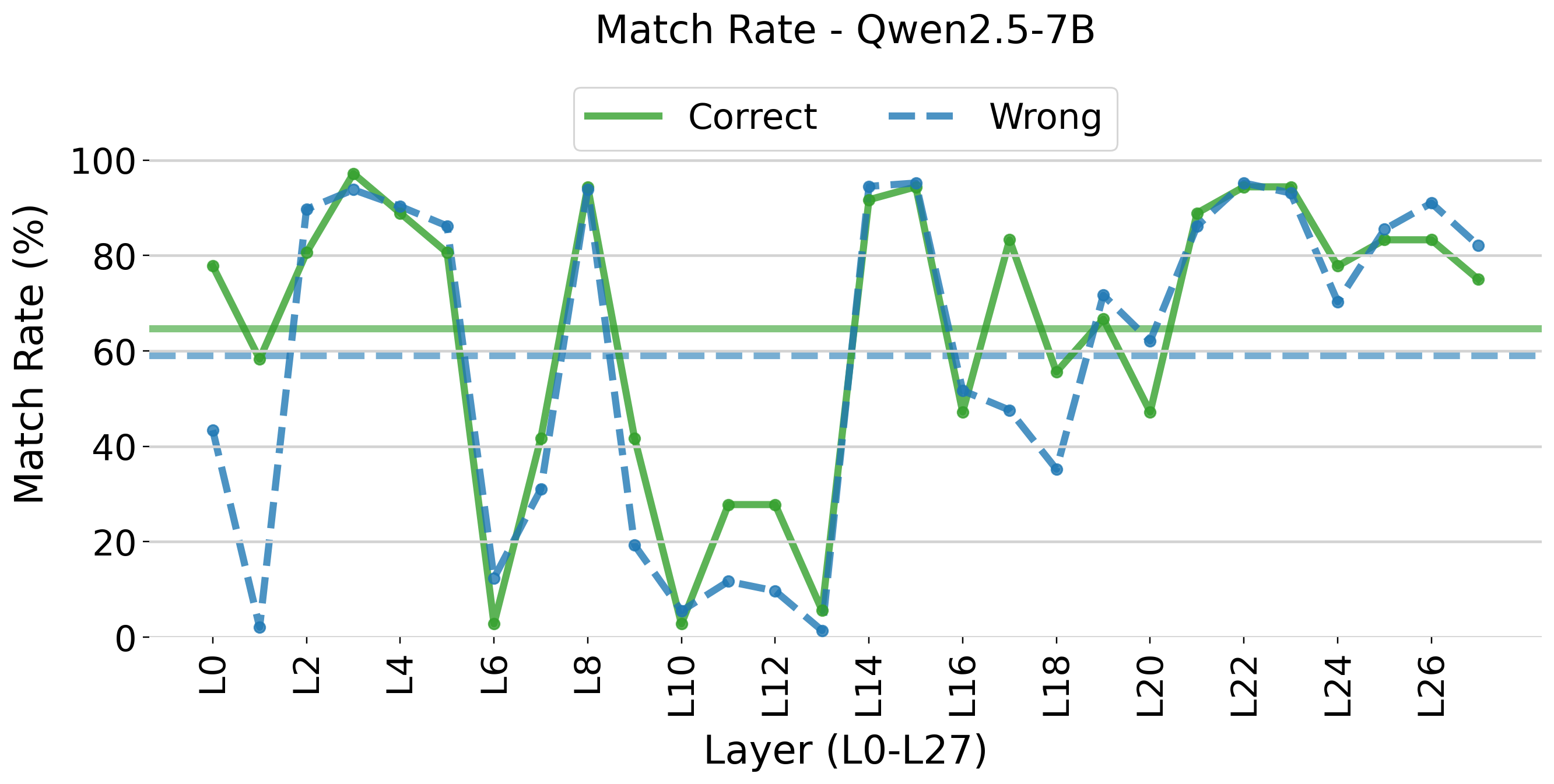}
        \caption{Match rate for Qwen2.5-7B.}
        \label{fig:8a}
    \end{subfigure}
    \vspace{-0.4em}
    \begin{subfigure}{0.85\linewidth}
        \centering
        \includegraphics[width=\linewidth]{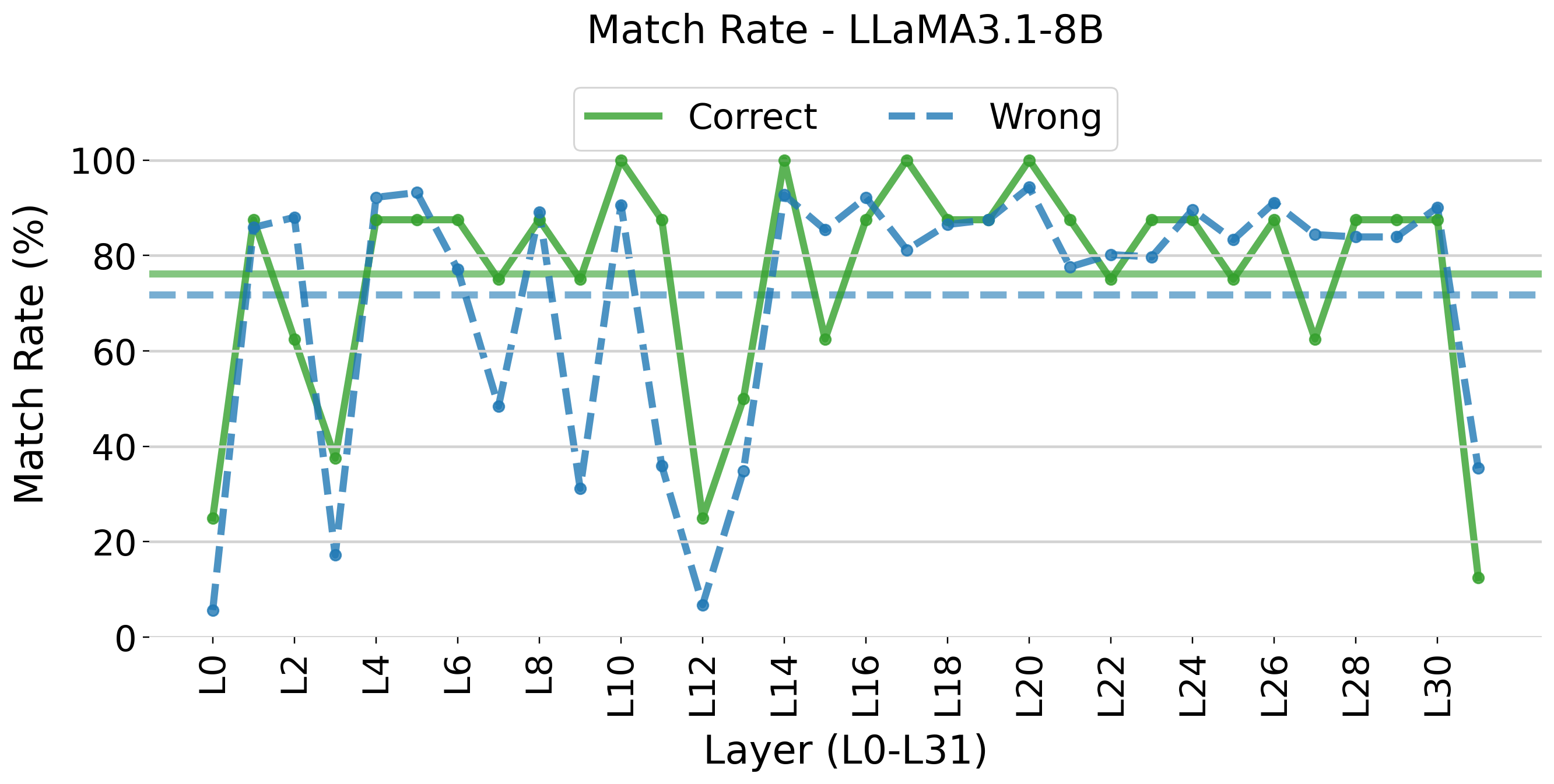}
        \caption{Match rate for LLaMA3.1-8B.}
        \label{fig:8b}
    \end{subfigure}

    \caption{Match rate across layers (x-axis). }
    \label{fig:8}
\end{figure}

\noindent \textbf{2. Layer-wise attention scores may lock onto the final-output candidate at shallow layers.} Figure~\ref{fig:8} reports the match rate between the top attention candidate per layer and the final output, probing whether attention converges with depth and aligns with the eventual decision. Results show pronounced layer-to-layer match rate fluctuations with depth, and becomes more stable in deeper layers. Appendix~\ref{Appendix:E.1} (Figure~\ref{fig:combo_attn_alignment}) further confirms substantial cross-depth match rate variation. High match rates in shallow layers suggest that LLMs may commit to their eventual decision early (making this metric a potential early signal of answer reliability), while deep-layer attention reallocation, likely for complex information integration, drives the non-monotonic trend.



\begin{figure}[htbp]
    \centering
    \begin{subfigure}[b]{0.48\linewidth}
        \centering
        \includegraphics[width=\linewidth]{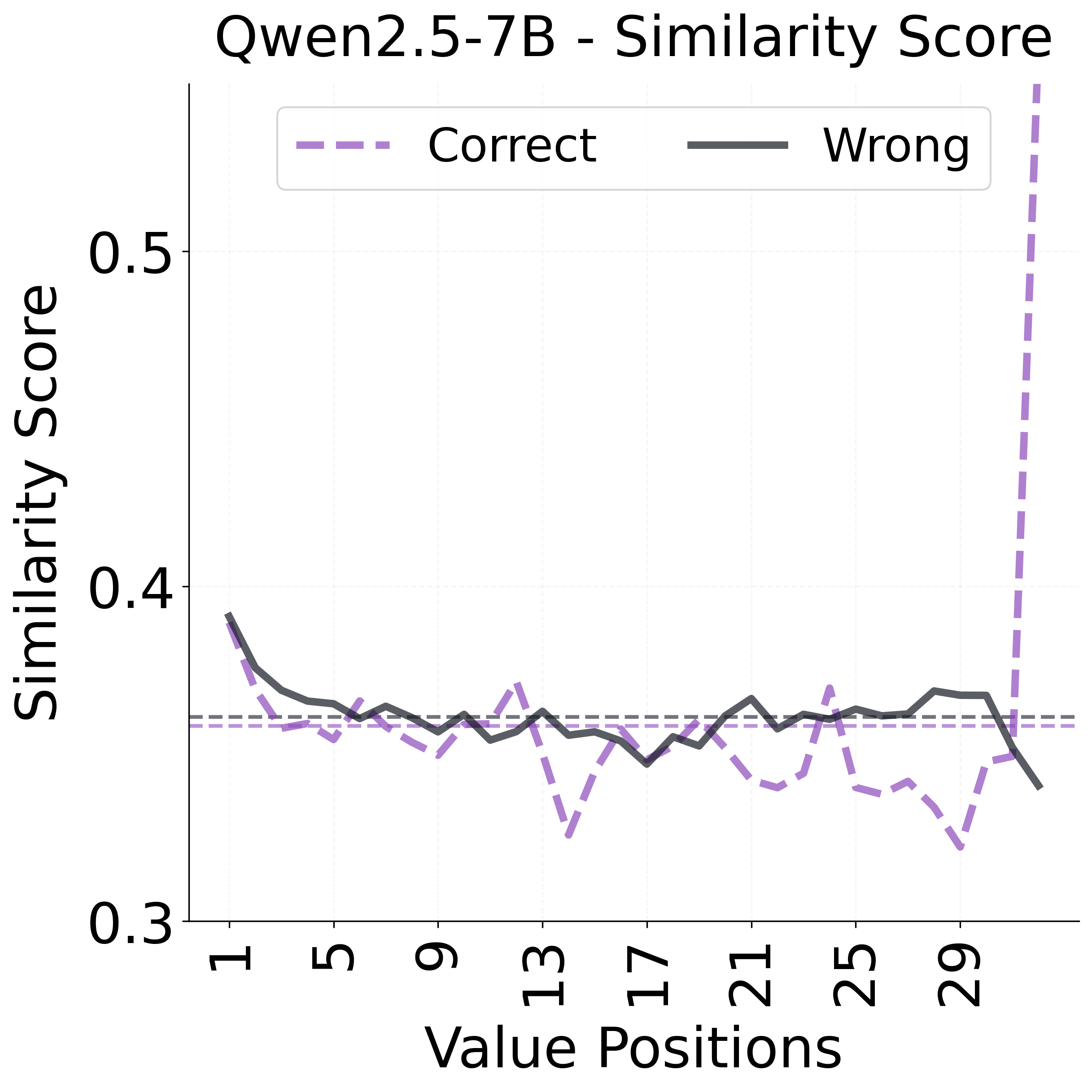}
        \caption{Qwen2.5-7B.}
        \label{fig:11a}
    \end{subfigure}
    \hfill
    \begin{subfigure}[b]{0.48\linewidth}
        \centering
        \includegraphics[width=\linewidth]{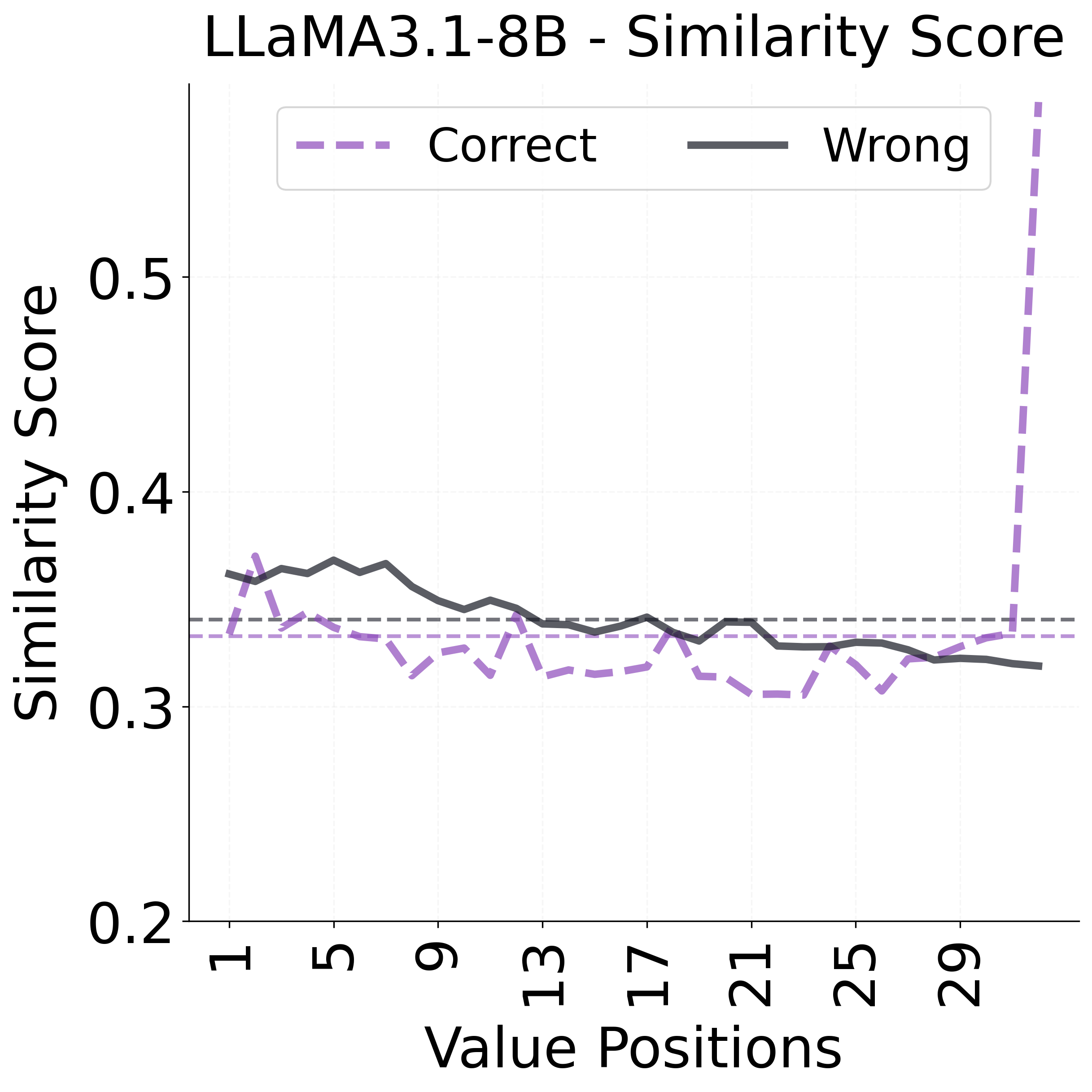}
        \caption{LLaMA3.1-8B.}
        \label{fig:11b}
    \end{subfigure}
    \caption{Hidden-state similarity to candidates(x-axis).}
    \label{fig:11}
    \vspace{-5pt}
\end{figure}

\begin{table*}[t]
\centering
\setlength{\tabcolsep}{4.5pt}
\renewcommand{\arraystretch}{0.9}
\resizebox{\textwidth}{!}{%
\begin{tabular}{ll c c c c c c c c}
\toprule
\multirow{2}{*}{\textbf{Model}} & \multirow{2}{*}{\textbf{Query}} & \multirow{2}{*}{\textbf{WO Intervention}}
& \multicolumn{3}{c}{\textbf{General Prompting}}
& \multicolumn{2}{c}{\textbf{General Mnemonic}}
& \multicolumn{2}{c}{\textbf{Memory Updating}} \\
\cmidrule(lr){4-6}\cmidrule(lr){7-8}\cmidrule(lr){9-10}
& & & CoT & 2-Shot & Index & Rehearsal & Semantic & Integration & Forgetting \\
\midrule
\multirow{3}{*}{\textbf{LLaMA3.1-8B}}
& Earliest & 96.34 & 95.73\downm & \textcolor[HTML]{104680}{\textbf{97.56}}\upm & 95.12\downm & 95.12\downm & 95.12\downm & \textcolor[HTML]{8A2230}{\textbf{96.95}}\upm & 95.73\downm \\
& Latest   & 75.61 & 80.49\upm   & \textcolor[HTML]{104680}{\textbf{84.76}}\upm & 79.88\upm   & 76.83\upm   & 76.22\upm   & \textcolor[HTML]{8A2230}{\textbf{84.76}}\upm   & 76.83\upm \\
& ELAG      & 20.73 & 15.24\downm & 12.80\downm & 15.24\downm & 18.29\downm & 18.90\downm & 12.19\downm & 18.90\downm \\
\midrule
\multirow{3}{*}{\textbf{Qwen2.5-7B}}
& Earliest & 99.39 & \textcolor[HTML]{104680}{\textbf{100.00}}\upm & 98.78\downm & 96.95\downm & 99.39\      & 100.00\upm & 99.39\       & \textcolor[HTML]{8A2230}{\textbf{100.00}}\upm \\
& Latest   & 88.41 & 87.80\downm & \textcolor[HTML]{104680}{\textbf{89.02}}\upm   & 85.98\downm & 87.20\downm & 87.20\downm & \textcolor[HTML]{8A2230}{\textbf{89.02}}\upm  & 87.80\downm \\
& ELAG      & 10.98 & 12.20\upm   & 9.76\downm  & 11.27\upm   & 12.19\upm   & 12.80\upm   & 10.37\downm & 12.20\upm \\
\bottomrule
\end{tabular}%
}
\caption{Performance of intervention methods on the real-world dataset (\%): accuracy on earliest/latest queries and the resulting gap. General mnemonic and memory updating strategies are two cognitively inspired method families (Rehearsal/Semantic vs.\ Integration/Forgetting). Blue text marks the best-performing method among general LLM prompting baselines, and red text marks the best-performing cognitively inspired method.}
\label{tab:earliest_latest_gap}
\end{table*}

\begin{figure}[htb]
\centering
\includegraphics[width=0.45\textwidth]{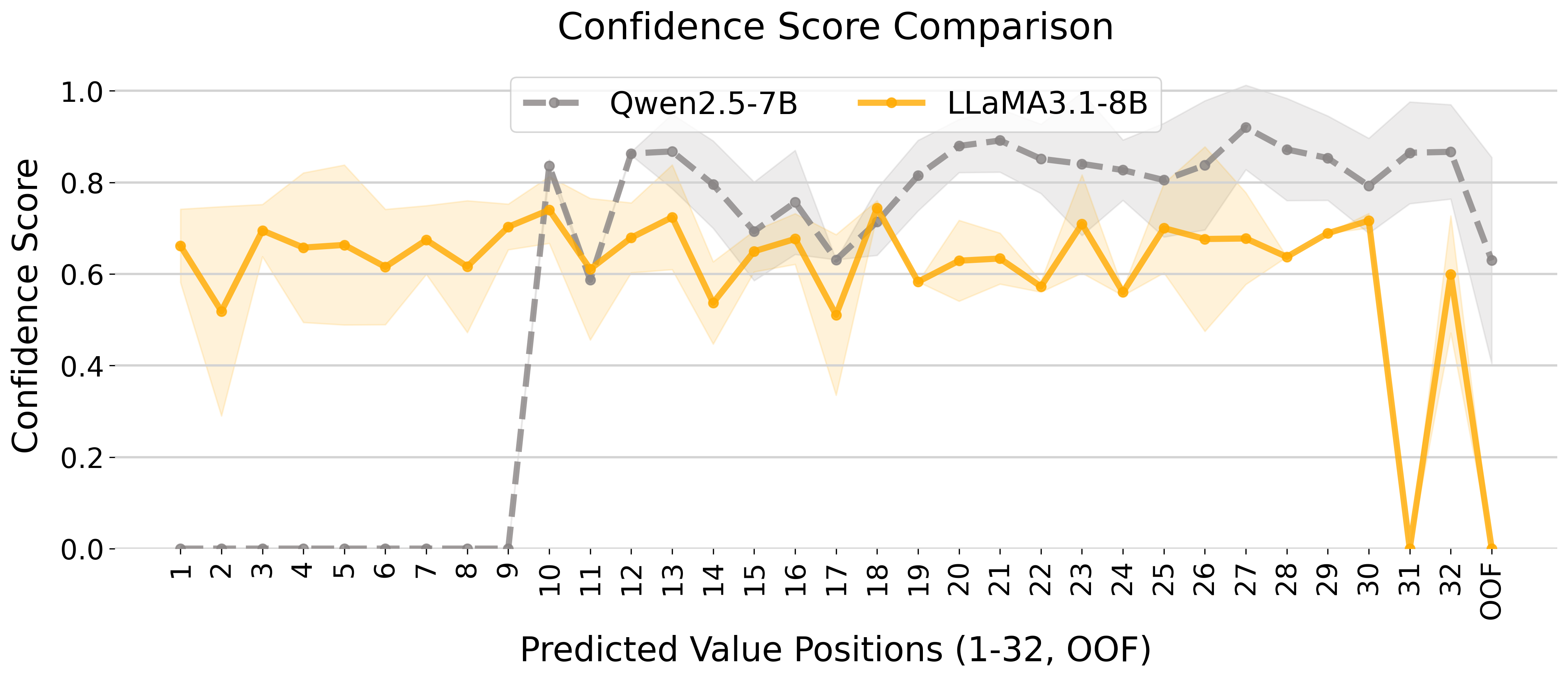}
\centering
\caption{Answer confidence distributions. For candidate positions(x-axis), samples without a corresponding predicted answer are assigned a confidence of 0.}
\label{fig:13}
\end{figure}

\noindent \textbf{3. Hidden-state similarity peaks on correct samples but flatter with weaker peaks on errors.} Figure~\ref{fig:11} shows the similarity distributions aggregated across samples and layers (Eq.~\ref{eq:7}), testing whether the representation at the answer position can geometrically align with the correct candidate and separate it from competing alternatives. The results indicate that when the model fails, the answer-position representation does not form stable and separable evidence to support selecting the latest update. Appendix~\ref{Appendix:E.2} (Figure~\ref{fig:12}) further reports and analyzes the layer-wise evolution of the similarity scores.



\noindent \textbf{4. LLMs exhibit divergent uncertainty expression patterns, and logits lack stable peaks on errors.} Figure~\ref{fig:13} presents confidence scores (Eq.~\ref{eq:9}): Qwen2.5-7B typically outputs high confidence and often remains confident even when incorrect, suggesting overconfidence, whereas LLaMA3.1-8B shows lower overall confidence and a more conservative response style. This suggests that using confidence as a proxy for uncertainty (e.g., threshold-based abstention) may be highly model-dependent in this task. Appendix~\ref{Appendix:E.3} (Figure~\ref{fig:14}) further analyzes candidate-wise logit distributions, showing that wrong cases lack stable logits peaks across latest positions.

Synthesizing three internal signal analyses, we conclude that when failing in latest-state retrieval, LLMs' evidence chain for the latest candidate systematically collapses, characterized by flattened signals, unstable peaks and inconsistent cross-layer alignment failing to provide reliable local evidence for locking onto the latest update. Such failures stem not from single-module/layer issues but from cross-layer, representation-to-decision evidence instability, with models unable to secure a decisive advantage for the latest value amid multi-candidate competition.


\vspace{-5pt}
\subsection{Heuristic Interventions}
\label{sec:4.6}

Given that local internal signals do not reveal an stable and effective angle for internal intervention, we introduce cognitively motivated heuristic interventions that use prompting to strengthen the model's internal encoding of the latest update or to reduce interference from competition among multiple candidate values. We compare these heuristic interventions with common LLM prompting strategies (CoT, 2-shot, and Index), with detailed prompt designs in Appendix~\ref{Appendix:D.1}. Table~\ref{tab:earliest_latest_gap} reports the intervention results on the real-world dataset for Qwen2.5-7B and LLaMA3.1-8B. The results demonstrate that two interventions, 2-shot prompting and memory integration, can improve latest-state retrieval while preserving earliest-state accuracy, outperforming methods based on additional reasoning steps or simple rehearsal. However, they still cannot fully eliminate the models' retrieval bias. Appendix~\ref{Appendix:D.2} (Table~\ref{tab:rw_interventions_mean_std}) further analyzes the intervention results on the synthetic dataset. Future work should develop more targeted model-side update-tracking mechanisms to fundamentally mitigate the retrieval bias.



\vspace{-5pt}
\section{Conclusion}
We propose the DKI evaluation framework and use it to examine how LLMs retrieve knowledge when the same fact is updated multiple times within context. Across experiments, we observe a pronounced retrieval bias: earliest-state accuracy stays high, while latest-state accuracy drops. Internal-signal diagnostics show that, in error cases, these signals become flatter and less discriminative, offering little stable local evidence to identify the correct update. Our cognitively inspired prompt interventions can partially improve the latest state retrieval, but they do not fully eliminate the bias. Future work should center on knowledge-update tracking and develop targeted methods to fundamentally mitigate the bias.


\section{Limitations}
First, our evaluation adopts endpoint probing, querying only the earliest and latest states. This setup enables a low-ambiguity and high-controllability assessment of multi-update retrieval performance, but it narrows observable behavioral scope: we cannot directly measure the model's ability to track and retrospectively retrieve intermediate update states. Future work could preserve the core setting of multiple updates to the same fact while introducing higher-level task abstractions beyond the cue-value format (e.g., structured assessments such as interval judgments, ordering comparisons, or reconstruction of state-transition structures), to more systematically evaluate a model's capability to localize and recover intermediate historical states.

Second, although we conduct experiments on both synthetic and real-world data, there remains an ecological gap between the two. Synthetic DKI construction weakens semantic information and parametric priors, potentially limiting the generalizability of our conclusions to complex, semantically diverse real-world knowledge-updating scenarios. Future work should build richer long-form real-world datasets with multiple updates, covering longer updates and  broader domains, to more reliably assess LLM performance and failure boundaries in practice.

Third, our internal-signal analyses mainly provide correlational diagnostics, and they cover only a limited set of model families, precluding strong causal claims about underlying mechanisms. Future work should take a more causally testable perspective, vis controlled interventions and ablation studies, to identify the key integrated internal signals driving update retrieval success or failure, and further explore robust model-side improvement directions based on these findings.





\bibliography{custom}

\appendix

\section{Experiment Setup}
\label{Appendix:A}

Table~\ref{tab:1} summarizes the configurations of our synthetic and real-world dynamic knowledge instances, the full model suite considered in our study, and the hyperparameters. 
For each model, we also list an abbreviated identifier in parentheses; for brevity, these short names are used throughout the main text and figures.

\begin{table*}[htbp]
\centering
\begin{tabular}{p{0.35\textwidth}p{0.6\textwidth}}
\toprule
\textbf{Category} & \textbf{Configuration and description} \\
\midrule
\multicolumn{2}{l}{\textbf{Data (dynamic knowledge instances)}} \\
Synthetic data size & 200 synthetic DKIs \\
Synthetic data updates $T$ & $\{32, 64, 128, 256, 512\}$ \\
Real-world data size & 164 dynamically updated DKIs \\
Real-world data updates $T$ & Average $8.77$, minimum $2$, maximum $70$ \\
Temperature & 0.0 \\
Random Seed & $\{0, 1, 2, 3, 4\}$ \\
Evaluation metrics & Earliest Accuracy, Latest Accuracy, and Earliest - Latest Accuracy Gap ($Acc_{earliest} - Acc_{latest}$), reported as mean over seeds\\
\midrule
\multicolumn{2}{l}{\textbf{Model suite}} \\
Model families & Mainstream commercial and open-source LLMs (LLaMA~3.1, Qwen2.5/3, GPT, Claude, Gemini, DeepSeek) \\
Primary experimental models & \texttt{llama-3.1-405b-Instruct} (LLaMA3.1-405B), \texttt{llama-3.1-70b-Instruct} (LLaMA3.1-70B), \texttt{llama-3.1-8b-Instruct} (LLaMA3.1-8B) \\
 & \texttt{Qwen3-235B-A22B-Instruct-2507} (Qwen3-235B), \texttt{Qwen2.5-72B-Instruct} (Qwen2.5-72B), \texttt{Qwen2.5-32B-Instruct} (Qwen2.5-32B), \texttt{Qwen2.5-14B-Instruct} (Qwen2.5-14B), \texttt{Qwen2.5-7B-Instruct} (Qwen2.5-7B) \\
API models & \texttt{gpt-5-nano} (GPT5-N), \texttt{gpt-4.1-nano} (GPT4.1-N), \texttt{claude-sonnet-4-20250514} (Claude-S-4), \texttt{gemini-2.5-flash-preview-05-20} (Gemini2.5-F), \texttt{DeepSeek\_V3} (DeepSeek-V3) \\
Internal signal analysis models & \texttt{Qwen2.5-7B-Instruct} (Qwen2.5-7B), \texttt{Llama-3.1-8B-Instruct} (LLaMA3.1-8B) \\
Qwen2.5-7B & 28 attention heads and 28 layers \\
LLaMA3.1-8B & 32 attention heads and 32 layers \\
\bottomrule
\end{tabular}
\caption{Overview of experimental settings and model suite. Abbreviated model names in parentheses are used in figures and analysis.}
\label{tab:1}
\end{table*}

\begin{figure*}[htb]
\centering
\includegraphics[width=0.99\textwidth]{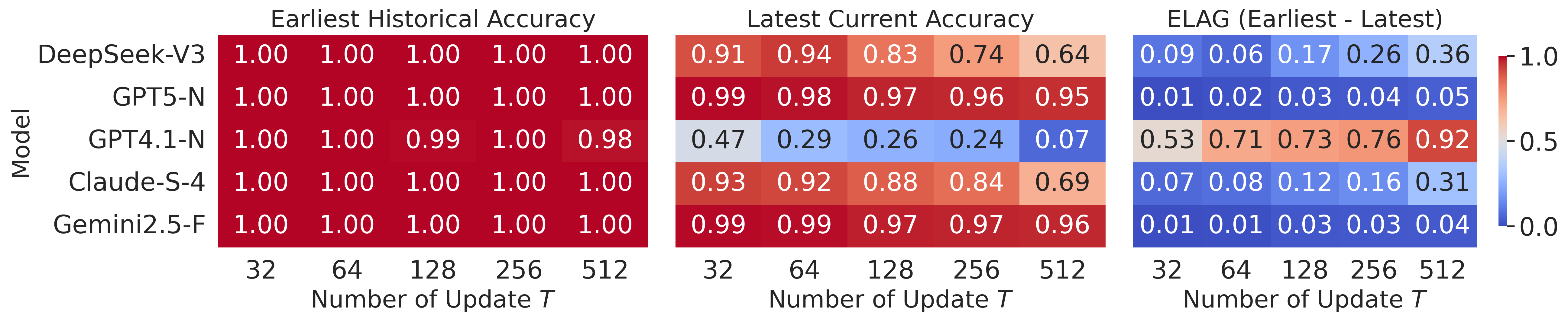}
\centering
\caption{The performance of different LLMs in terms of the number of updates $T$ for dynamic knowledge instances.}
\label{fig:3}
\end{figure*}

\begin{figure*}[h]
    \centering
    
    \begin{subfigure}[b]{0.89\textwidth}
        \centering
        \includegraphics[width=\textwidth]{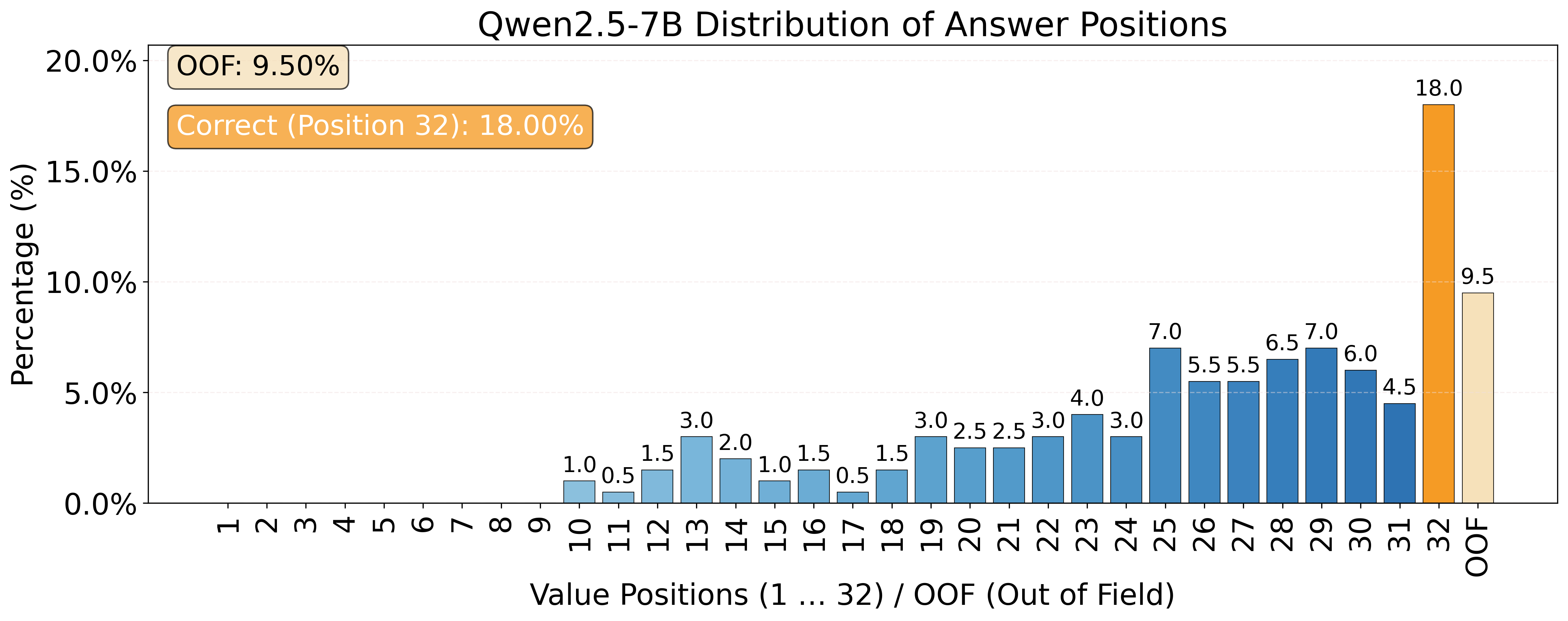}
        \caption{ Qwen2.5-7B exhibits strong bias toward late update positions with non-trivial out-of-field outputs.}
        \label{fig:7a}
    \end{subfigure}
    
    \begin{subfigure}[b]{0.89\textwidth}
        \centering
        \includegraphics[width=\textwidth]{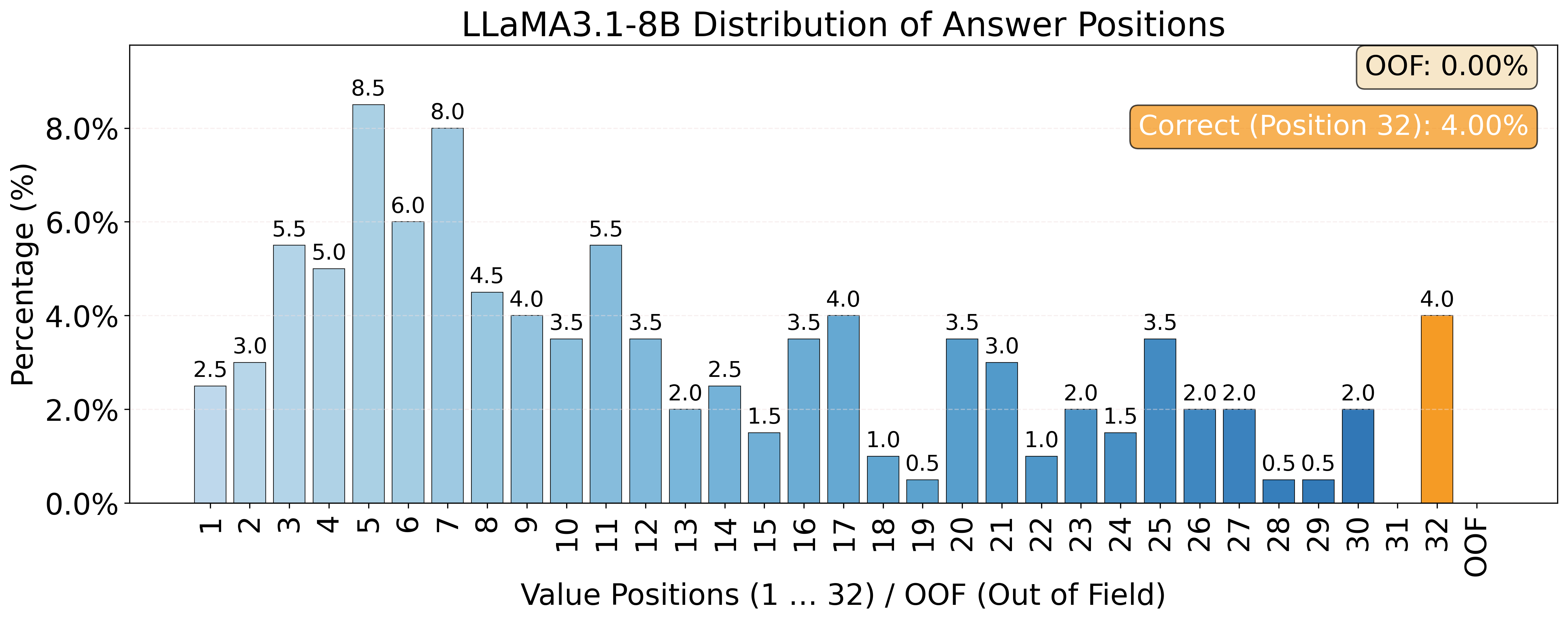}
        \caption{LLaMA3.1-8B shows a more dispersed distribution biased toward earlier updates.}
        \label{fig:7b}
    \end{subfigure}
    
    \caption{Predicted answer-position distribution for latest-state queries ($T=32$).}
    \label{fig:7}
\end{figure*}

\section{DKI Prompt Example}
\label{Appendix:B}

We present the concrete prompt design in the DKI Prompt Example below. This design continuously updates the value associated with the same cue to simulate the ongoing evolution of knowledge, and by comparing the model's responses to the earliest historical information and the latest current state, we assess whether its observable behavior exhibits retrieval bias.

\vspace{6pt}
\vspace{6pt}
\begin{tcolorbox}[
    colback=gray!2,
    arc=5pt,
    left=10pt,
    right=10pt,
    top=6pt,
    bottom=6pt,
    title= DKI Prompt Example,
    fonttitle=\large,
    center title,
    breakable,
]

You are given a long updated list of cue-value records and a target cue (CUE).\\
For this target cue, return its earliest historical and latest current VALUE according to FIRST and LAST occurrence order in the provided record list.\\[6pt]

\textbf{CUE (JSON array):} \texttt{["President of Italy"]}\\[8pt]

\textbf{INPUT FORMAT}\\[-3pt]

- Each record is one line in the form: \texttt{cue:value}

- Boundaries: lines strictly between the literal markers \texttt{START:} and \texttt{END}

\textbf{Record List}\\[-3pt]

\texttt{START:}\\
President of Italy:Alcide De Gasperi\\
President of Italy:Enrico de Nicola\\
President of Italy:Luigi Einaudi\\
President of Italy:Giovanni Gronchi\\
President of Italy:Antonio Segni\\
President of Italy:Giuseppe Saragat\\
President of Italy:Giovanni Leone\\
President of Italy:Sandro Pertini\\
President of Italy:Francesco Cossiga\\
President of Italy:Oscar Luigi Scalfaro\\
President of Italy:Carlo Azeglio Ciampi\\
President of Italy:Giorgio Napolitano\\
President of Italy:Sergio Mattarella\\
\texttt{END}\\[8pt]

\textbf{Output (valid JSON only):}\\[-3pt]
\texttt{\{"cue":"<cue>", "earliest":"<VERBATIM or UNKNOWN>","latest":"<VERBATIM or UNKNOWN>"\}},\\[8pt]

\textbf{Rules:}
\begin{itemize}
    \item Only the cue specified by CUE.
    \item Earliest/Latest are defined strictly by first/last appearance order within \texttt{START..END}.
    \item Keep VALUE VERBATIM; JSON-escape only as required.
    \item Output exactly one JSON object and nothing else. No code, no prose, no markdown.
\end{itemize}

\end{tcolorbox}

\begin{figure}[htb]
\centering
\includegraphics[width=0.49\textwidth]{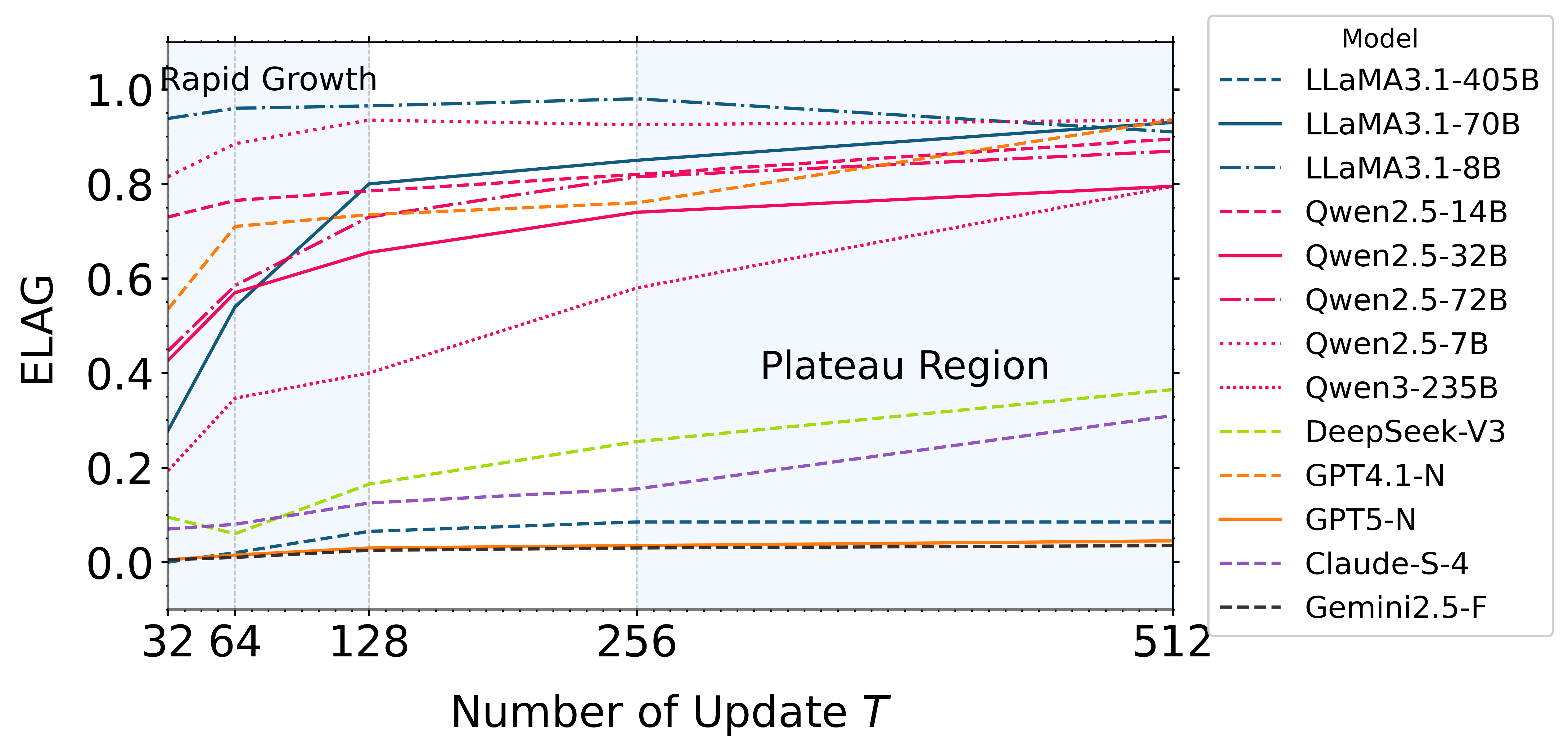}
\centering
\caption{ELAG as a function of update length $T$ on synthetic DKIs. Each curve corresponds to one model, and models are grouped by family. Lower ELAG values indicate reduced asymmetry between earliest- and latest-state retrieval, reflecting improved resistance to update interference.}
\label{fig:4}
\end{figure}

\section{DKI Experimental Results}

\subsection{Retrieval Bias Phenomenon Identification}

\label{Appendix:C.1}

\noindent \textbf{Performance of Other LLMs.} To further assess generality, Figure \ref{fig:3} reports results on five additional commercial/API models (DeepSeek-V3, GPT5-N, GPT4.1-N, Claude-S-4, and Gemini2.5-F), which exhibit the same trend: near-perfect accuracy on the earliest historical state across all $T$, but degraded accuracy on the latest current state as $T$ increases, leading to pronounced retrieval bias.

\noindent \textbf{Ealiest-Latest Gap Dynamics.} To summarize retrieval behavior under multiple in-context updates, we analyze the earliest-latest accuracy gap as a function of update length $T$. For each model and each $T$, we compute
\begin{equation}
\mathrm{ELAG}(T) = \mathrm{Acc}_{\text{earliest}}(T) - \mathrm{Acc}_{\text{latest}}(T),
\end{equation}
where $\mathrm{Acc}_{\text{earliest}}$ is accuracy on earliest-state queries (targeting $V^{(1)}$)
and $\mathrm{Acc}_{\text{latest}}$ is accuracy on latest-state queries (targeting $V^{(T)}$).
A larger gap indicates a stronger divergence between retrieving the initial versus the most recent state,
and we use it as an operational measure of retrieval bias under multiple in-context updates.

Figure~\ref{fig:4} plots $\mathrm{ELAG}(T)$ for all $13$ evaluated LLMs on synthetic DKIs.
Across model families, we observe a consistent two pattern. First, the gap increases rapidly for $T\in[32,128]$, indicating that interference effects emerge quickly as updates increases.
Second, for $T\ge256$, many models enter a plateau regime where the gap saturates. Smaller and mid-sized LLaMA and Qwen models quickly reach very large retrival bias values (often above $0.6$), whereas larger or more robust LLMs such as LLaMA3.1-405B, GPT5-N, and Gemini2.5-F maintain much weaker retrival bias across all update lengths, with DeepSeek-V3 and Claude-S-4 lying in between.

\subsection{Real-World Data Validation}
\label{Appendix:C.2}
\begin{figure}[htb]
\centering
\includegraphics[width=0.49\textwidth]{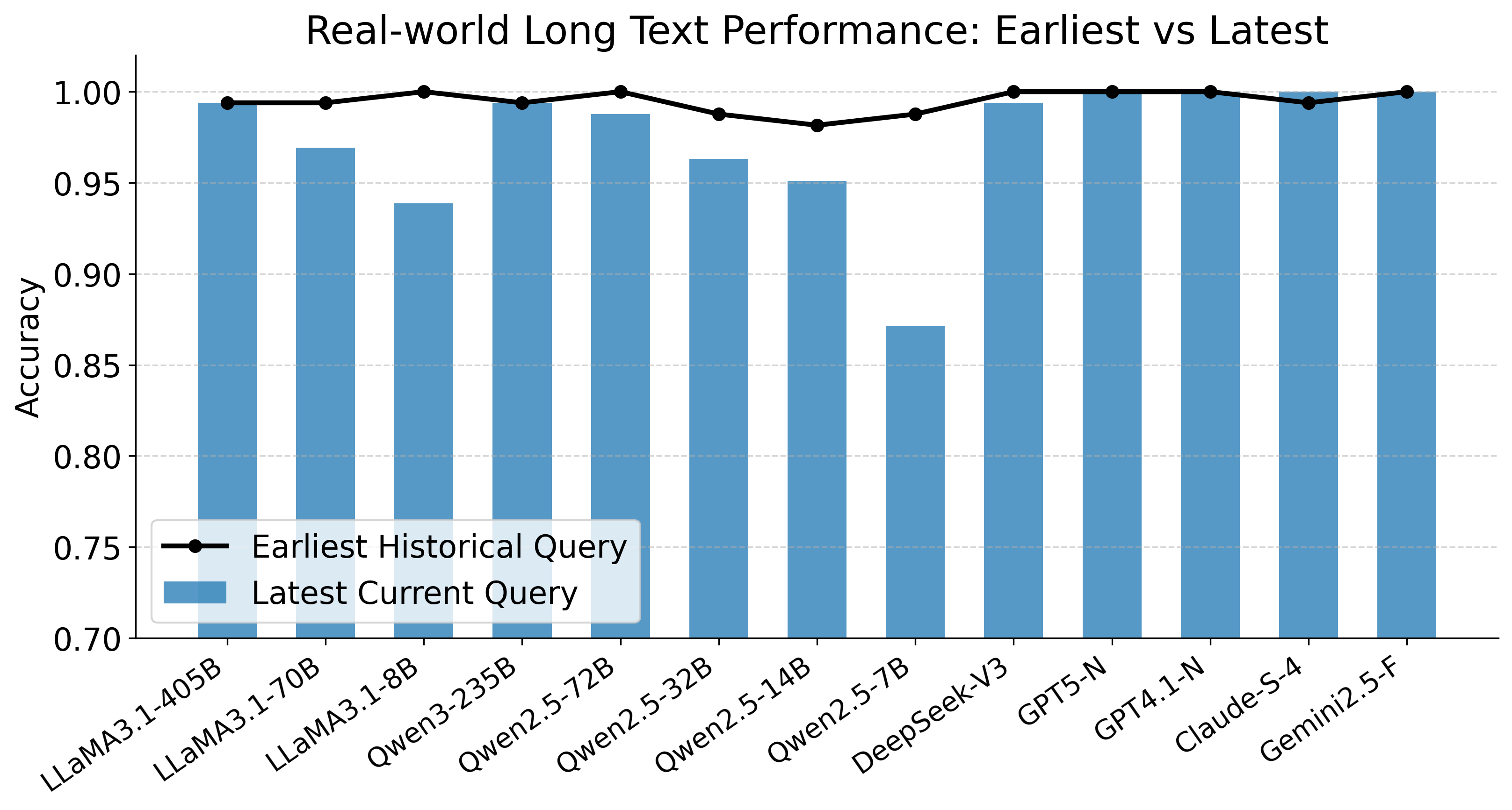}
\centering
\caption{Retrieval bias under a narrative long-text format (real-world DKIs).}
\label{fig:6}
\end{figure}

We further evaluate the same real-world DKI trajectories in a narrative long-text setting, where successive updates are integrated into a logically coherent document. This setup is more aligned with how users encounter dynamically evolving information in practical scenarios. Figure~\ref{fig:6} presents the corresponding experimental results. Below are examples of such long-text cases: these examples demonstrate a narrative-formatted dynamic knowledge instance, in which multiple temporally ordered updates are embedded in coherent narrative text, increasing the interference between the earlier and later states.

\begin{tcolorbox}[
    colback=gray!2,
    arc=5pt,
    left=10pt,
    right=10pt,
    top=6pt,
    bottom=6pt,
    title= Long-Text Example,
    fonttitle=\large,
    center title,
    breakable,
]


During the early post-war reconstruction period, the President of Italy was
Alcide De Gasperi, who played a pivotal role in stabilizing the country's
fragile democracy and fostering economic recovery after the devastation of the
Second World War. His leadership helped lay the groundwork for Italy's
integration into Western European institutions and the establishment of a new
republican constitution.

\medskip
Subsequently, we introduce, in the immediate aftermath of the war, the President of Italy was Enrico de
Nicola, who served as a transitional figure guiding Italy from monarchy to
republic. His presidency was marked by efforts to unify a nation deeply divided
by the legacy of fascism and war, overseeing the birth of the Italian Republic
and setting precedents for democratic governance.

\medskip
\ldots

Lastly, the current Italian President is Sergio Mattarella: acting as the head of state and guardian of the constitutional order, he emphasizes upholding the constitutional framework and national solidarity. Against the backdrop of growing political divisions and increasingly complex social issues, he has consistently served as an "institutional stabilizer", and safeguarded the continuity and legitimacy of the Italian Republic's institutional operation through the exercise of presidential powers. 
\end{tcolorbox}

\subsection{Error Characteristic Analysis}
\label{Appendix:C.3}

Figure~\ref{fig:7} illustrates, for latest-state queries, how model predictions distribute across candidate update positions. The results show that Qwen2.5-7B concentrates its predictions near the end of the update sequence while producing a non-trivial proportion of out-of-field (OOF) outputs. In contrast, LLaMA3.1-8B exhibits a more dispersed distribution that is overall biased toward earlier update positions, with almost no OOF outputs. These distributional patterns reveal different failure modes for latest-state retrieval, highlighting that the underlying mechanisms are strongly model-specific.

\section{Internal Signal Analysis}
\label{Appendix:E}

\subsection{Attention-based Scores}
\label{Appendix:E.1}
\begin{figure}[htbp]
    \centering

    \begin{subfigure}[t]{0.48\linewidth}
        \centering
        \includegraphics[width=\linewidth]{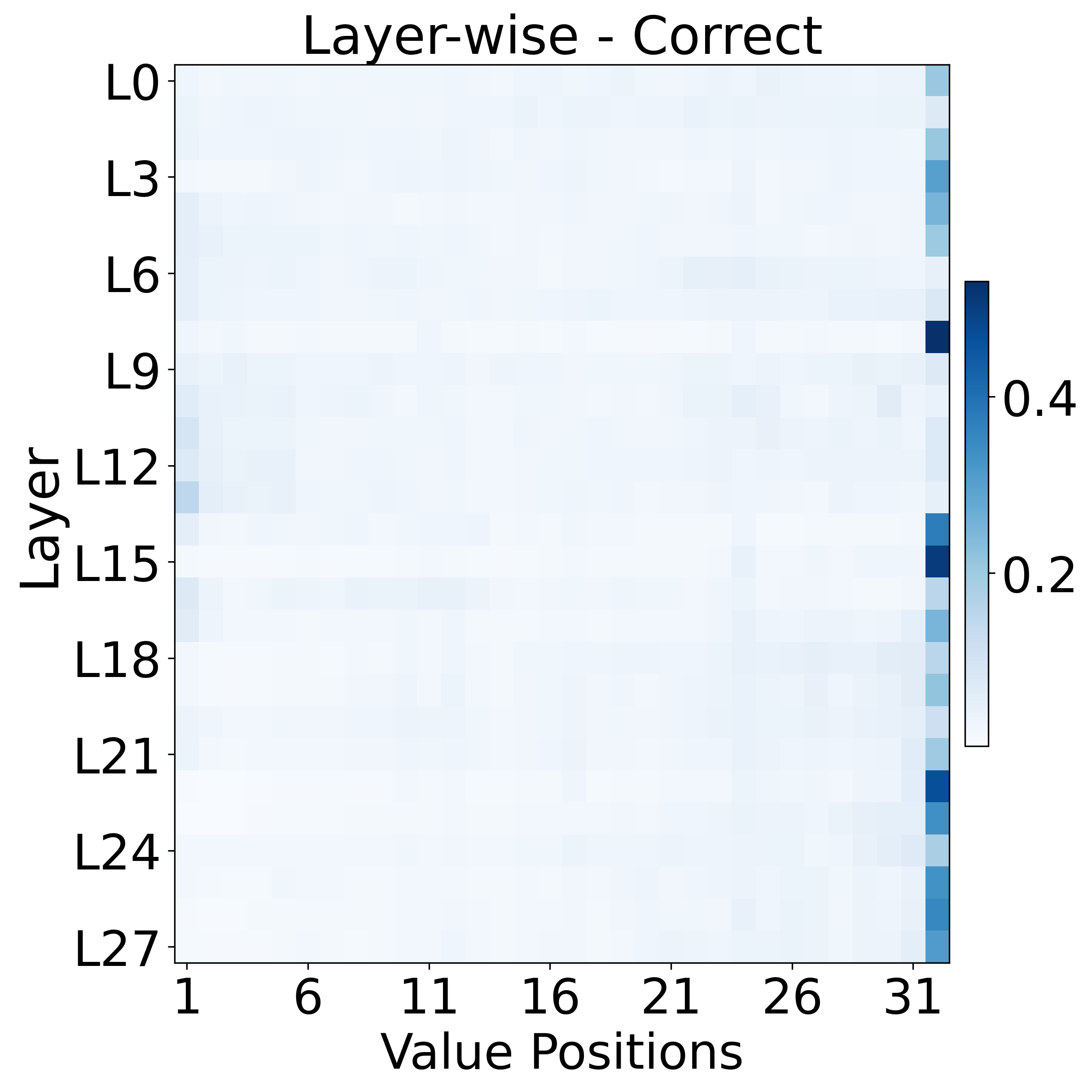}
        \caption{Qwen: Layer-wise.}
        \label{fig:att_correct_qwen_layer}
    \end{subfigure}
    \hfill
    \begin{subfigure}[t]{0.48\linewidth}
        \centering
        \includegraphics[width=\linewidth]{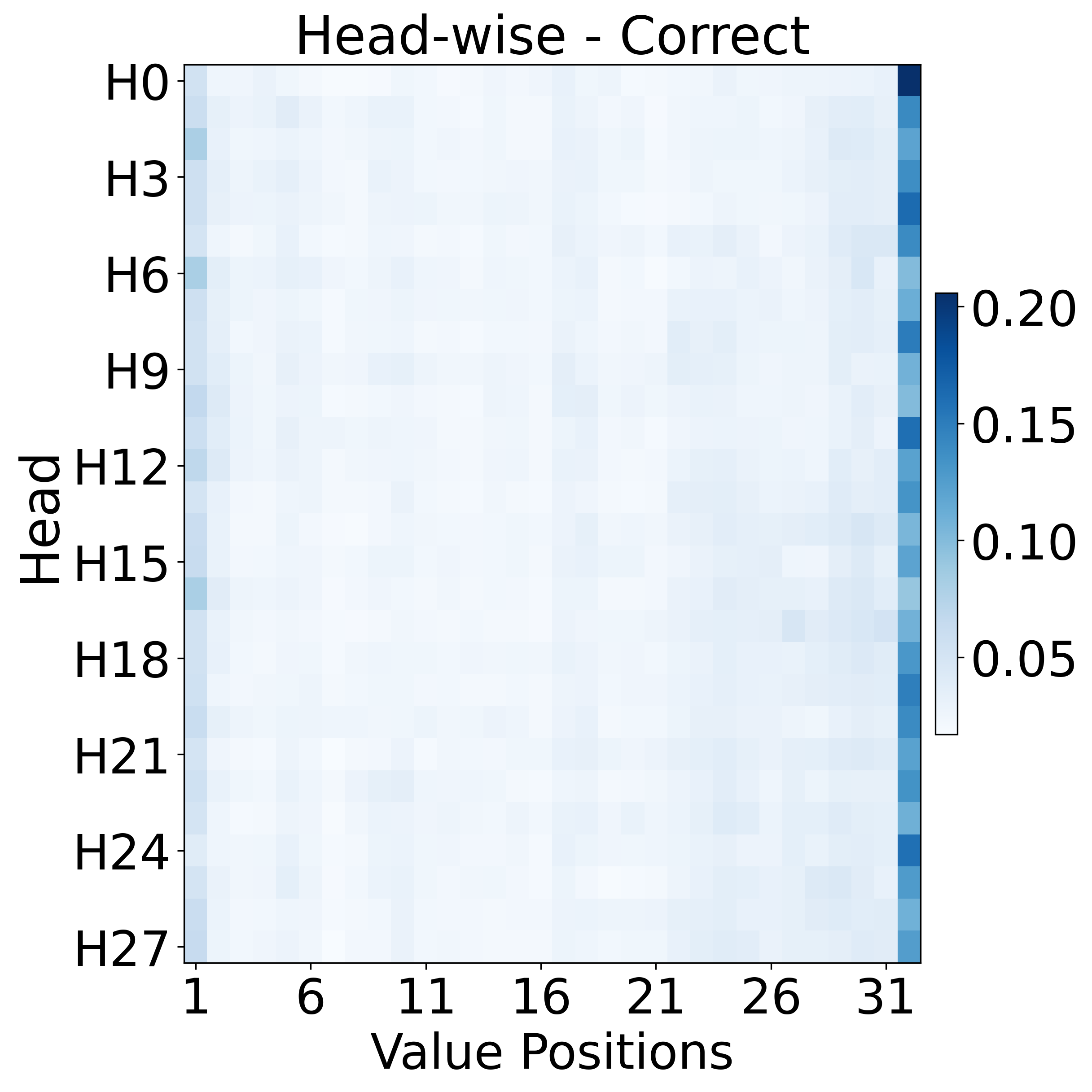}
        \caption{Qwen: Head-wise.}
        \label{fig:att_correct_qwen_head}
    \end{subfigure}

    \vspace{0.8em}

    \begin{subfigure}[t]{0.48\linewidth}
        \centering
        \includegraphics[width=\linewidth]{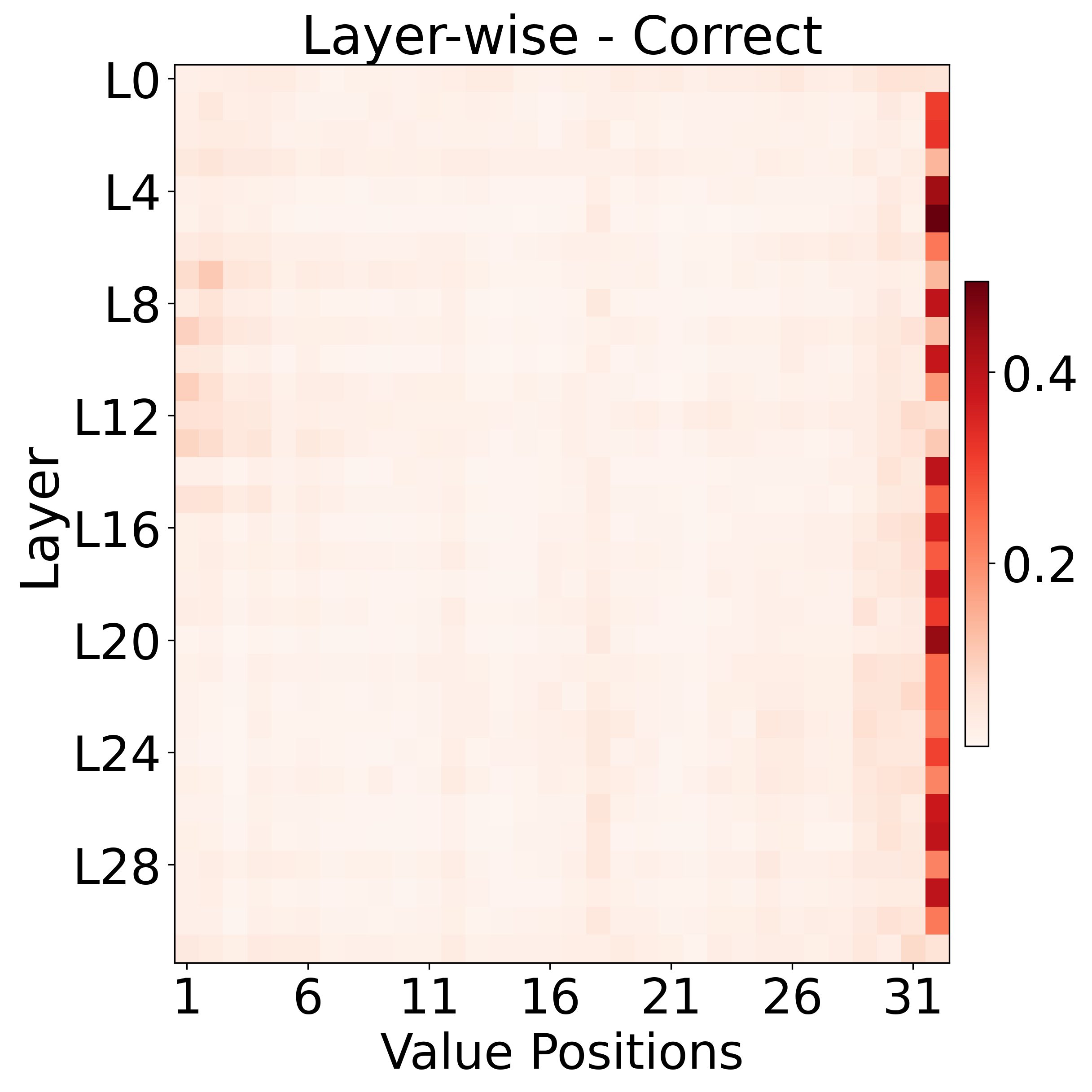}
        \caption{LLaMA: Layer-wise.}
        \label{fig:att_correct_llama_layer}
    \end{subfigure}
    \hfill
    \begin{subfigure}[t]{0.48\linewidth}
        \centering
        \includegraphics[width=\linewidth]{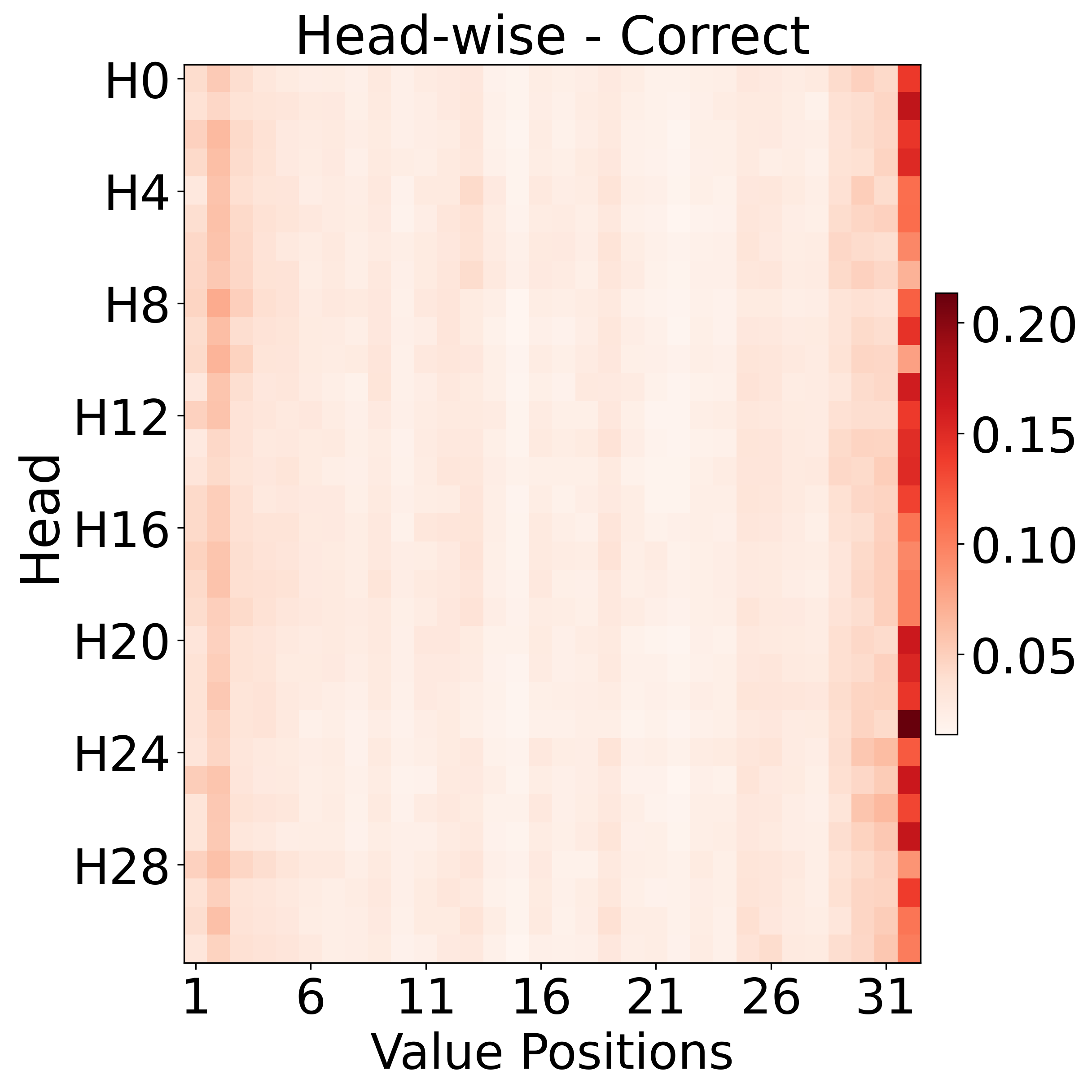}
        \caption{LLaMA: Head-wise.}
        \label{fig:att_correct_llama_head}
    \end{subfigure}

    \caption{Layer-wise and head-wise attention-score heatmaps over candidate positions (x-axis) on correct samples.}
    \label{fig:att_correct_all}
\end{figure}

\textbf{1. Layer-wise and Head-wise Attention-Score Heatmaps on Samples with Correct Predictions.} Figure \ref{fig:att_correct_all} presents the layer-wise and head-wise attention-score heatmaps of Qwen2.5-7B and LLaMA3.1-8B on samples with correct predictions. In both views, attention peaks at the position of the latest candidate value.

\begin{figure*}[htbp]
    \centering
    \begin{subfigure}[b]{0.85\textwidth}
        \centering
        \includegraphics[width=\linewidth]{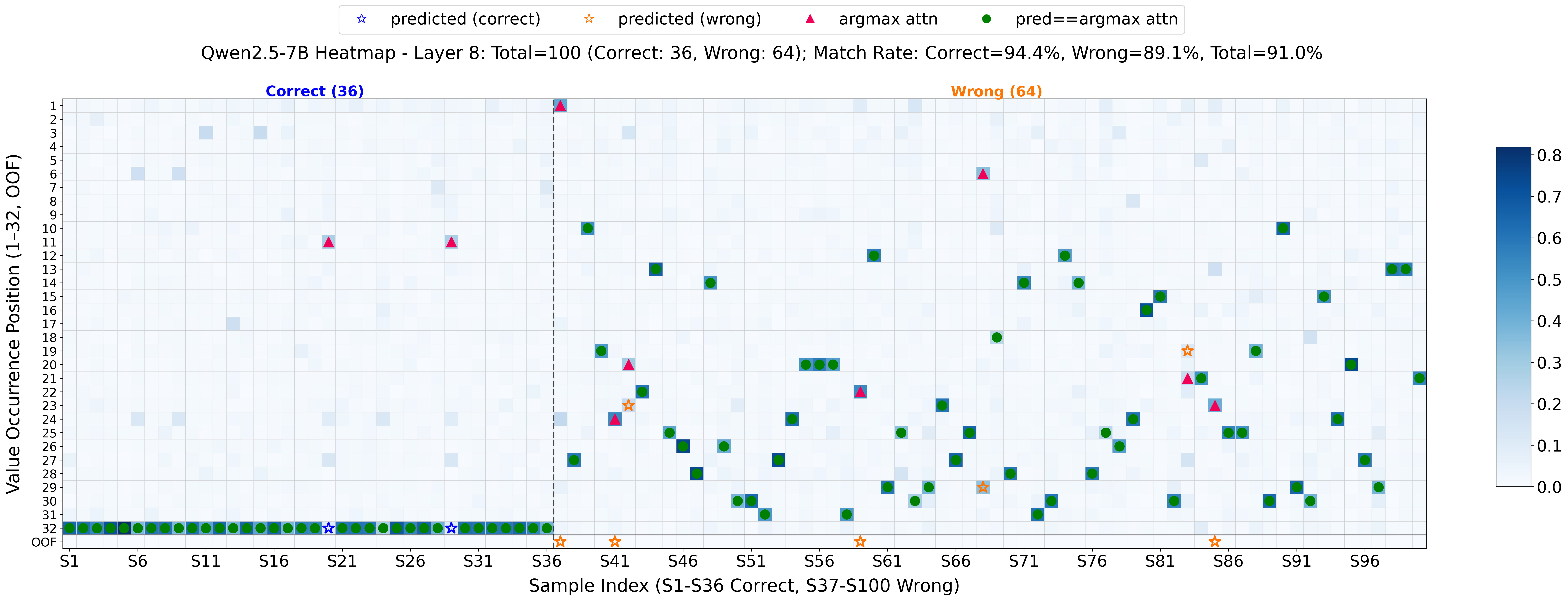}
        \label{fig:combo_attn_qwen_L08}
    \end{subfigure}
    \vspace{-3pt}
    \begin{subfigure}[b]{0.85\textwidth}
        \centering
        \includegraphics[width=\linewidth]{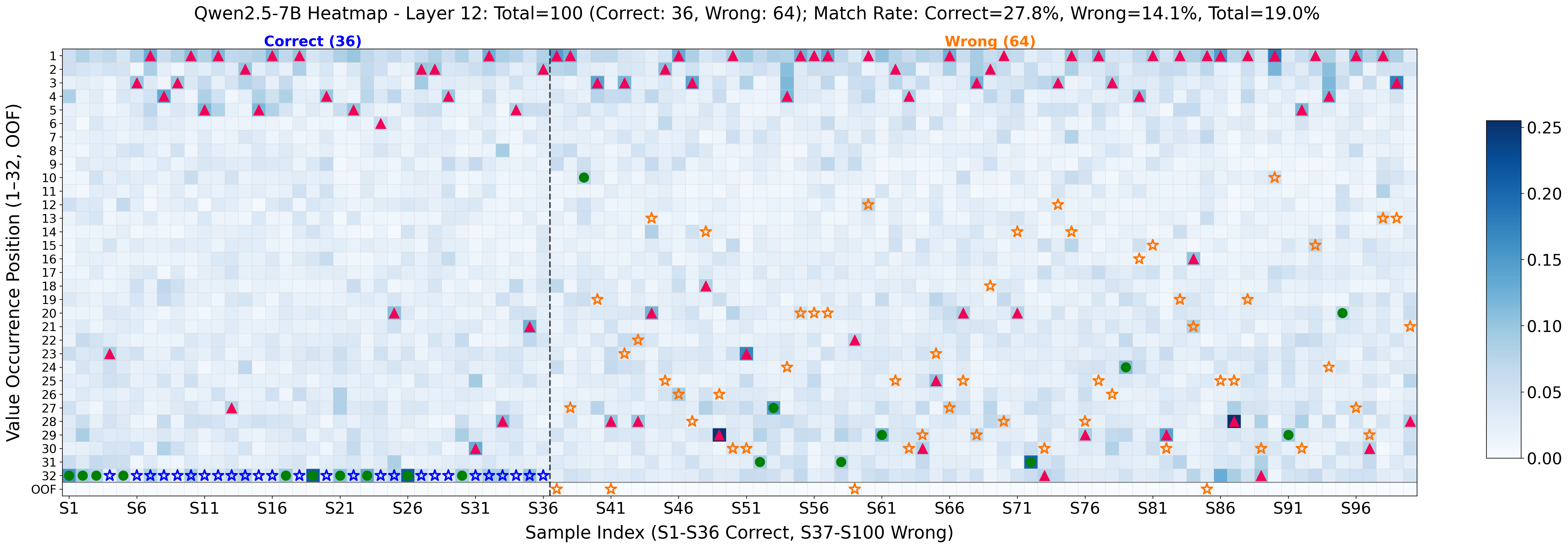}
        \label{fig:combo_attn_qwen_L12}
    \end{subfigure}
    \begin{subfigure}[b]{0.85\textwidth}
        \centering
        \includegraphics[width=\linewidth]{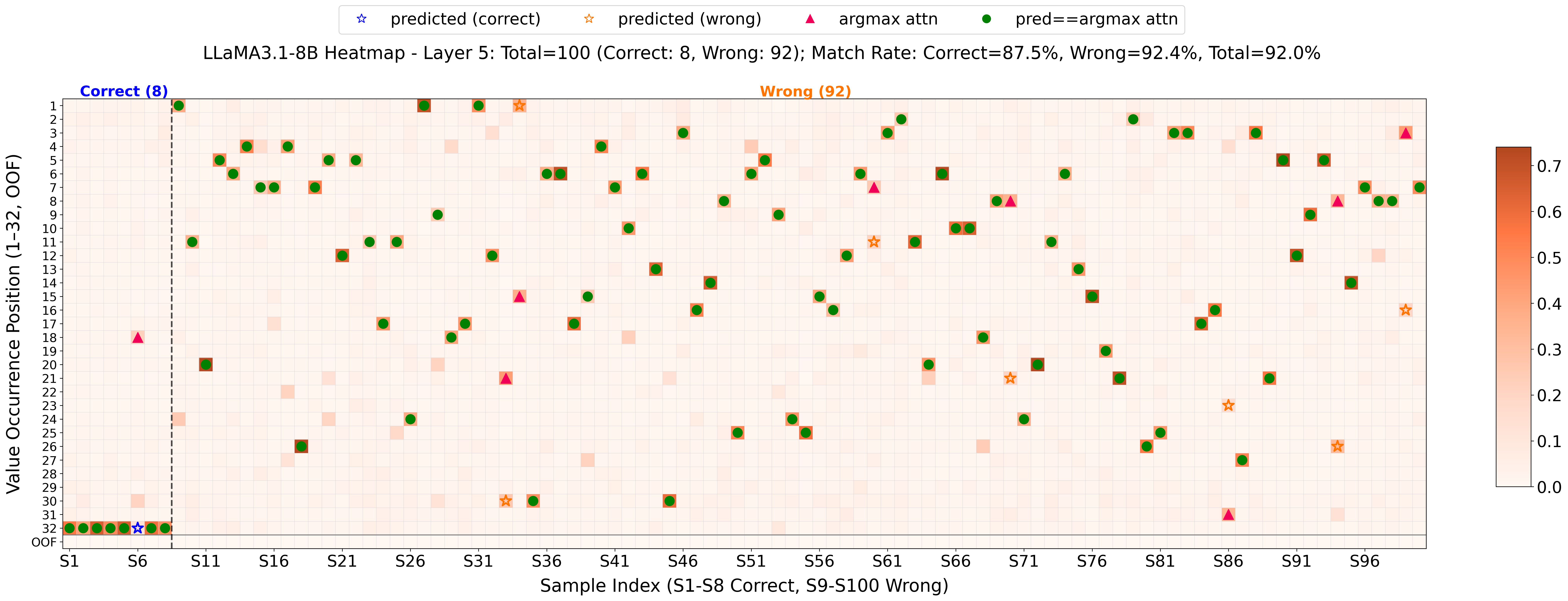}
        \label{fig:combo_attn_llama_L05}
    \end{subfigure}
    \vspace{-3pt}
    \begin{subfigure}[b]{0.85\textwidth}
        \centering
        \includegraphics[width=\linewidth]{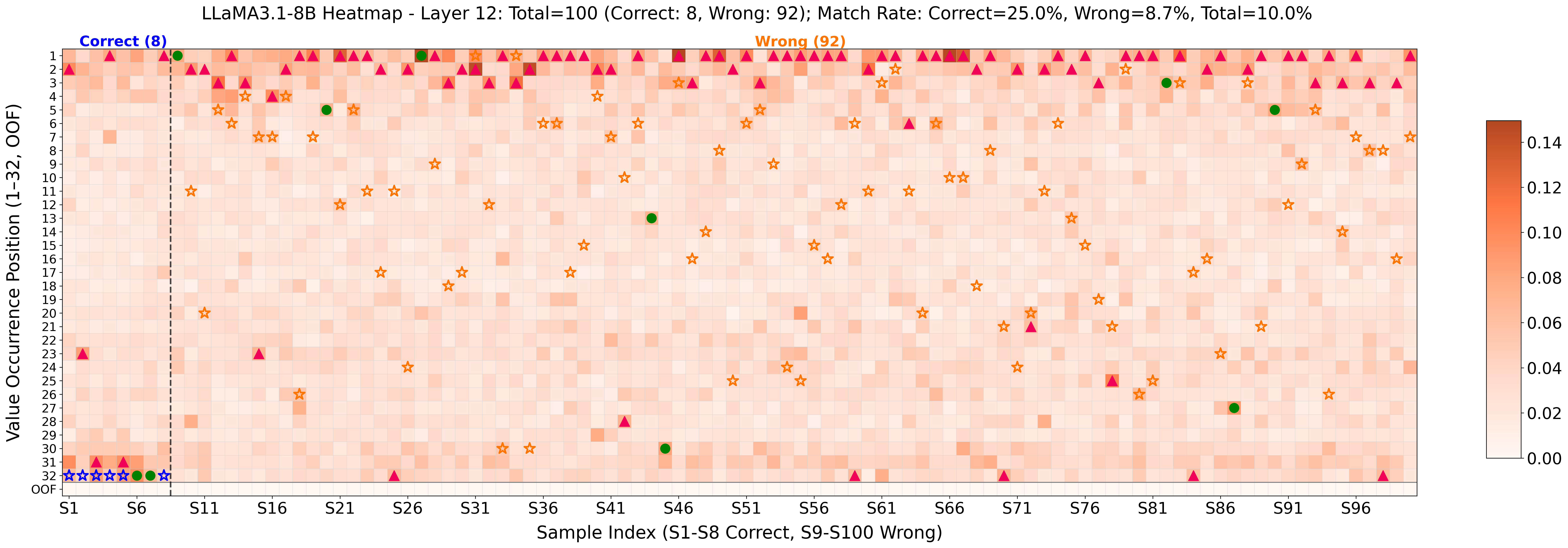}
        \label{fig:combo_attn_llama_L12}
    \end{subfigure}
    \caption{Comparing High- vs. Low-Match Layers for Attention-Output Alignment in Qwen2.5-7B and LLaMA3.1-8B.
The blue heatmaps correspond to Qwen2.5-7B, while the orange heatmaps correspond to LLaMA3.1-8B.
Correct samples are shown on the left and wrong samples on the right, separated by a dashed vertical line.
\HStar{blue} indicates the predicted candidate positions for correct samples, whereas \HStar{orange} indicates the predicted positions for wrong samples.
\STri{red} marks the candidate-value position with the highest attention score at the current layer;
\SDot{ForestGreen} indicates cases where this attention peak aligns with the model's final output (i.e., the two coincide).
}
    \label{fig:combo_attn_alignment}
\end{figure*}

\begin{figure*}[htbp]
    \centering
    
    \begin{subfigure}[b]{0.99\textwidth}
        \centering
        \includegraphics[width=\textwidth]{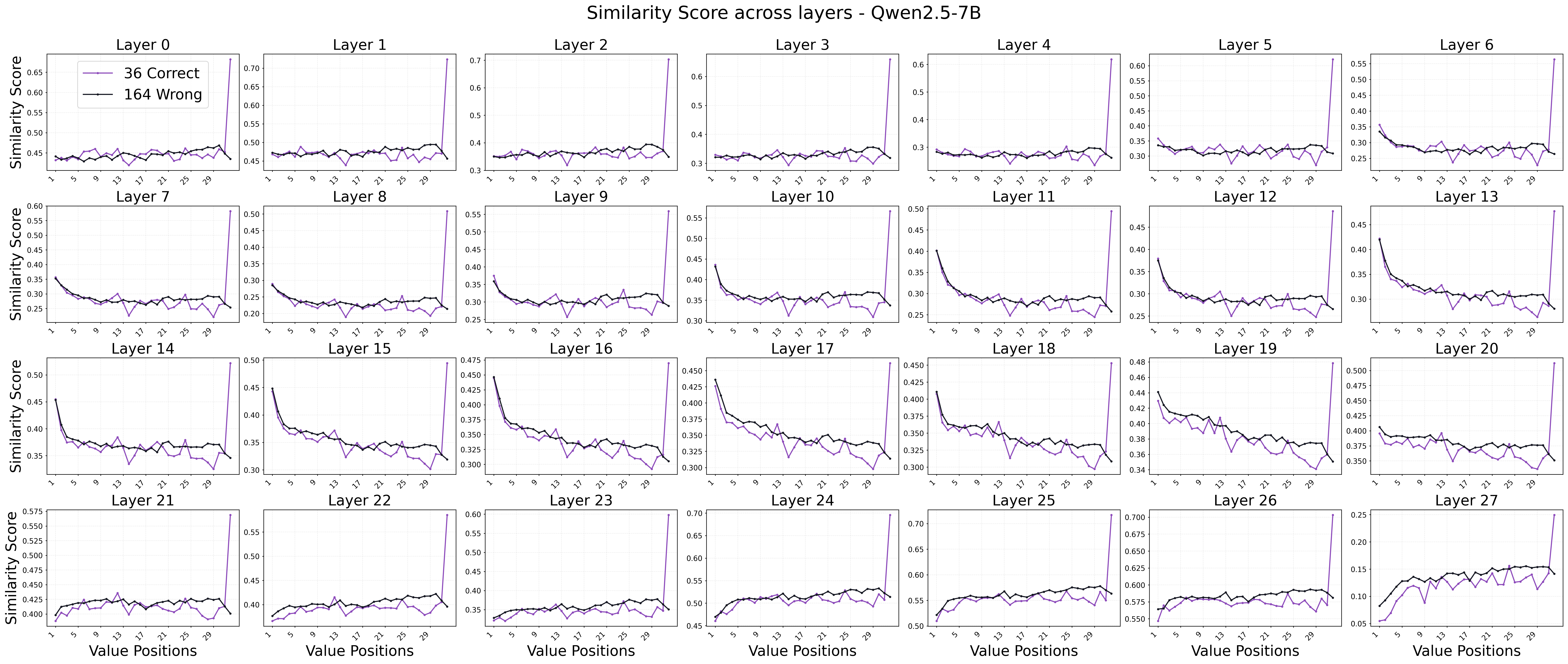}
        \caption{Similarity score at each layer for Qwen2.5-7B.}
        \label{fig:12a}
    \end{subfigure}
    
    \begin{subfigure}[b]{0.99\textwidth}
        \centering
        \includegraphics[width=\textwidth]{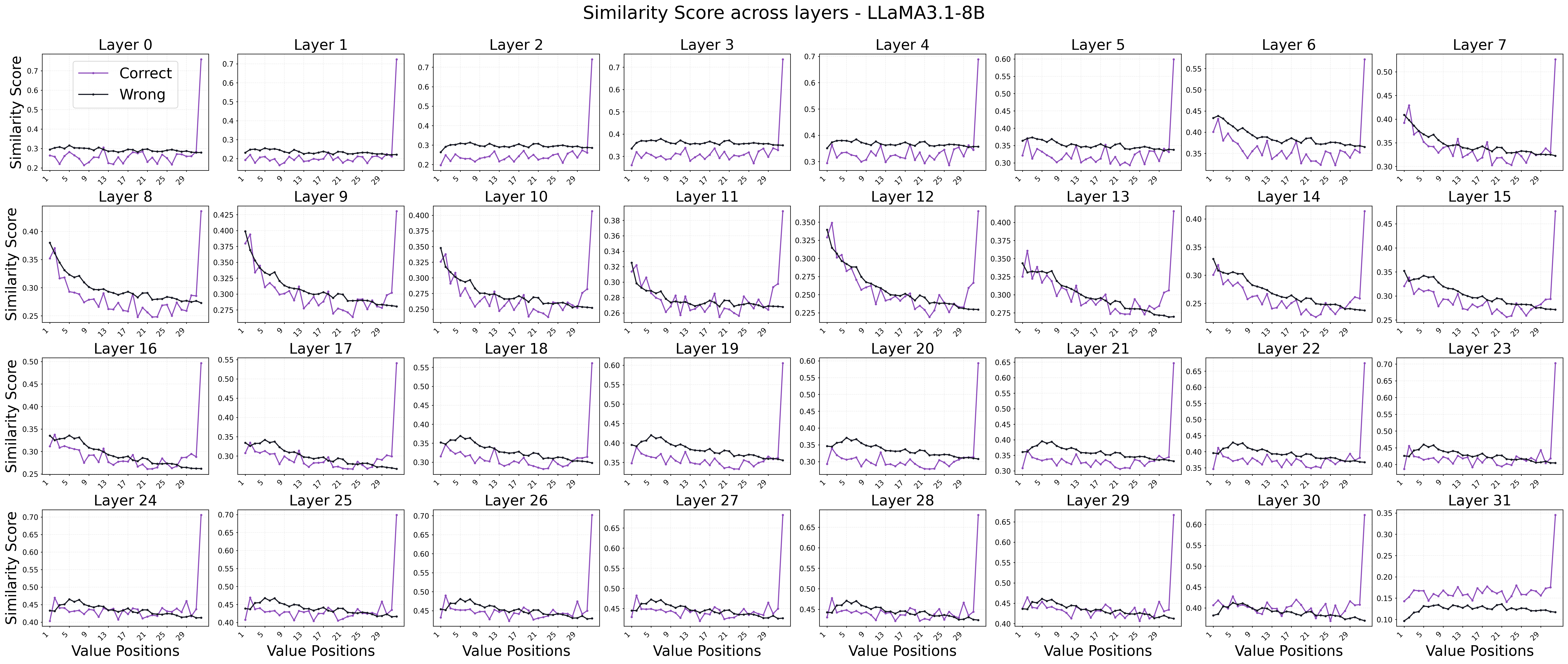}
        \caption{Similarity score at each layer for LLaMA3.1-8B.}
        \label{fig:12b}
    \end{subfigure}
    
    \caption{Layer-wise hidden-state similarity scores across candidate positions (x-axis). We report the mean similarity scores at each layer.}
    \label{fig:12}
\end{figure*}
\noindent \textbf{2. Layer-wise attention-output match rate analysis}. Figure~\ref{fig:combo_attn_alignment} presents sample-level visualizations to illustrate how the alignment between the candidate position receiving the peak attention at a given layer of the model and the model's final predicted answer position changes as the network layers deepen. For each model, we compare two representative layers: Layer 8 and Layer 12 for Qwen2.5-7B, and Layer 5 and Layer 12 for LLaMA3.1-8B. Qwen Layer 8 and LLaMA Layer 5 exhibit relatively higher attention-output match rates, whereas Qwen Layer 12 and LLaMA Layer 12 exhibit lower match rates.

Under the setting with a fixed update length $T=32$, we randomly sample 100 evaluation instances from each model for per-sample visualization. For Qwen2.5-7B, these include 36 correct and 64 wrong predictions; for LLaMA3.1-8B, due to the scarcity of correct cases overall, the 100 instances include 8 correct and 92 incorrect predictions. After comparing the two layers of each model, we observe a consistent pattern: in high-alignment layers, the coincidence frequency between attention peaks and the model's final outputs is higher (indicated by denser distributions of green circle markers); in contrast, in low-alignment layers, the deviation of attention peaks from the final outputs is more common (marked by more frequent separation between red triangle markers and pentagram markers), with the number of coincidences between them decreasing significantly.

These observations indicate that attention-output alignment does not follow a simple cumulative or monotonic trend with depth: higher alignment can emerge in relatively early layers, while deeper layers may still undergo shifts and reorganization in their attention distributions. In other words, some early layers already produce locally directed signals consistent with the final decision, but such evidence is not guaranteed to remain stable in later layers. The resulting layer-to-layer fluctuations therefore reflect functional differentiation of intermediate representations and attention mechanisms across depth, rather than a single deeper is more aligned trajectory.

\subsection{Hidden State Similarity Scores}
\label{Appendix:E.2}
To examine how the hidden-state similarity signal changes with depth, Figure~\ref{fig:12} visualizes layer-wise similarity scores for correct versus wrong samples. For Qwen2.5-7B, in early layers (L0 - L5), correct samples already exhibit a pronounced similarity peak at the latest candidate value, whereas the multi-sample average for wrong samples is relatively flat, indicating a lack of consistently directed representational evidence. In mid layers (L06 - L20), correct samples continue to maintain a sharp peak at the latest candidate; additionally, both correct and wrong samples show increased similarity toward early candidates. This may be because, in our endpoint-probing setting, the prompt requires predicting both the earliest and the latest values, so the model may transiently activate evidence associated with historical candidates in intermediate layers. In deeper layers (L21 - L27), the overall shape returns to one similar to L0 - L5, but the similarity curve exhibits an increasing slope toward later candidates, suggesting that the model tends to pull the answer representation toward the latest region in deeper layers, reflecting a form of coarse localization. LLaMA3.1-8B also shows a segmented, layer-dependent trend, but its deep-layer behavior differs from Qwen's: in deeper layers, similarity toward later candidates is harder to sustain and gradually weakens. This suggests that as depth increases, alignment of the answer representation to the latest candidate is not further consolidated; instead, the representation becomes more diffuse or drifts away, consistent with its overall weaker latest-state retrieval performance.

Overall, these results show that representational discriminability is not uniform across layers: correct samples preserve stable alignment to the latest candidate, while wrong samples display weaker and less consistent evidence, with Qwen retaining partial late-stage localization and LLaMA degrading more substantially with depth.

\subsection{Output Logits}
\label{Appendix:E.3}
\begin{figure*}[htb]
    \centering
    
    \begin{subfigure}[b]{0.99\textwidth}
        \centering
        \includegraphics[width=\textwidth]{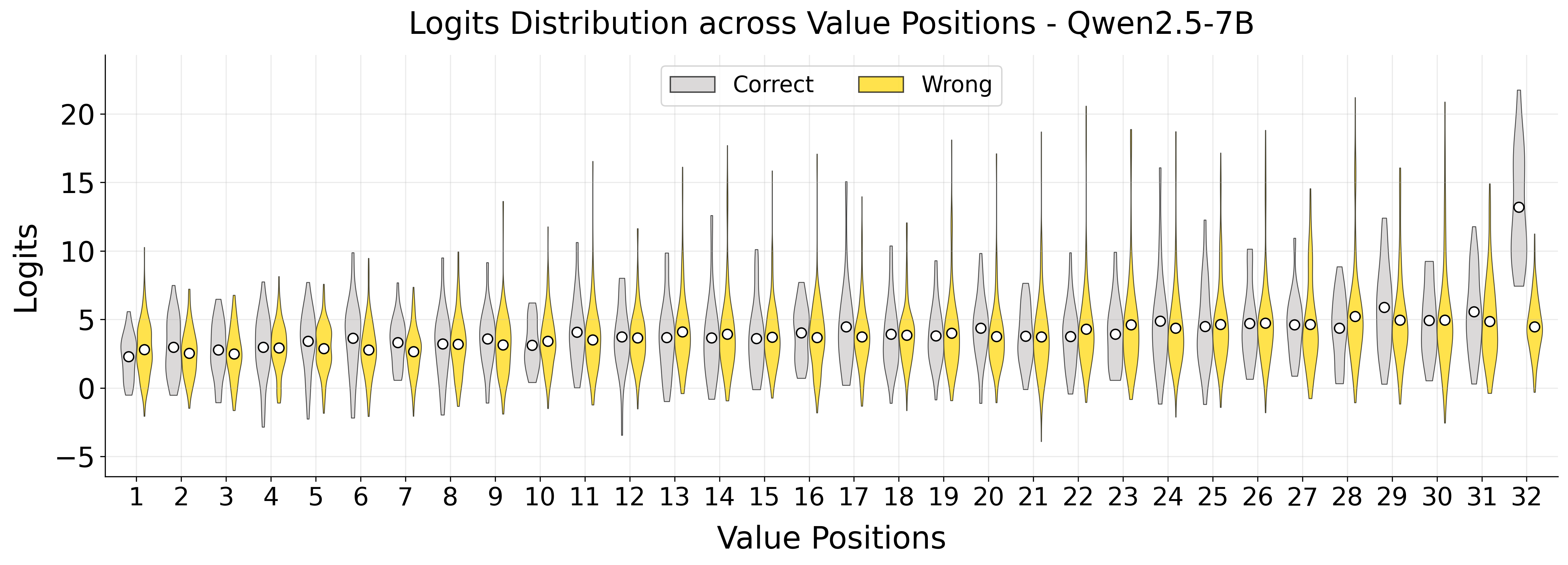}
        \caption{Qwen2.5-7B: output logits.}
        \label{fig:14a}
    \end{subfigure}
    
    \begin{subfigure}[b]{0.99\textwidth}
        \centering
        \includegraphics[width=\textwidth]{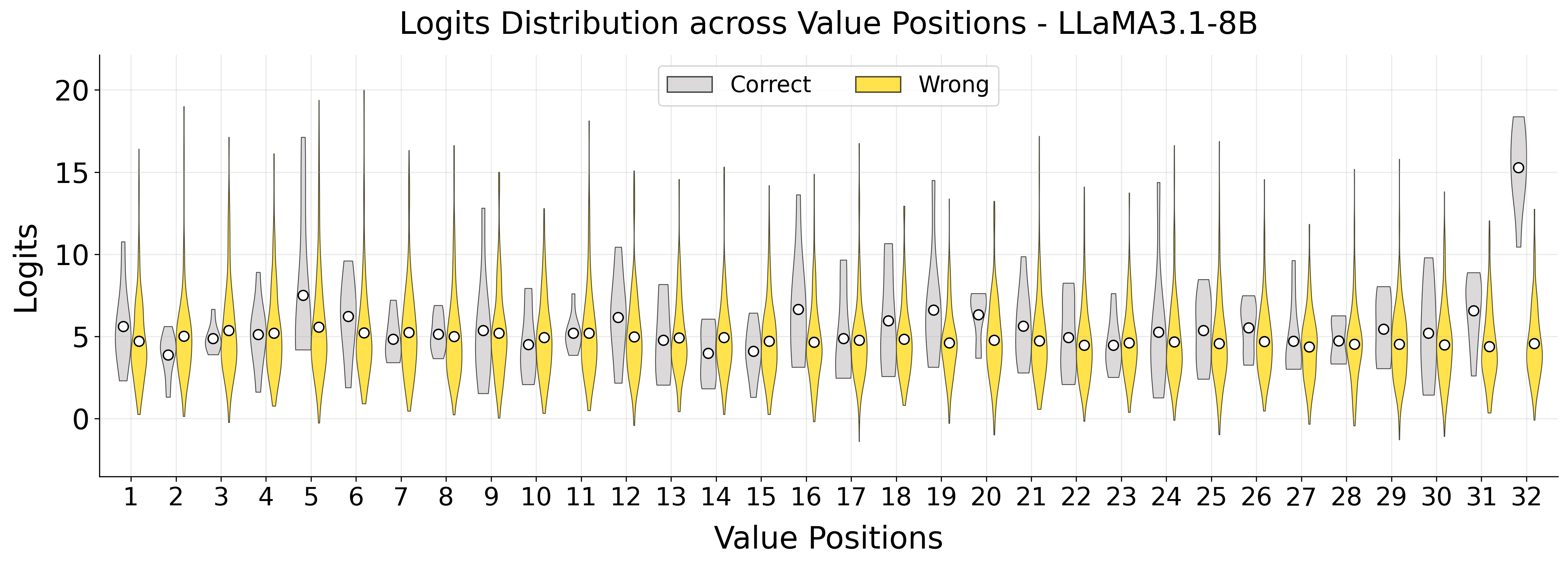}
        \caption{LLaMA3.1-8B: output logits.}
        \label{fig:14b}
    \end{subfigure}
    
    \caption{Output logit distributions over candidate answers(x-axis). Each violin shows the distribution of output logits for a given candidate, aggregated over samples, illustrating the variability of logits across candidates.}
    \label{fig:14}
\end{figure*}

Figure~\ref{fig:14} presents a comparison of logits across multiple candidate values for the LLMs. For correctly predicted samples from Qwen2.5-7B and LLaMA3.1-8B, the logits exhibit a distinct peak structure, where the logit corresponding to the correct candidate is significantly higher than those of the remaining candidates. For erroneously predicted samples, the correct candidate no longer maintains a stable dominance; instead, the logit distribution across all candidates becomes much flatter and lacks a prominent global peak. This indicates that when errors occur, the logit advantage is dispersed across different candidates rather than concentrating on a fixed incorrect candidate or the latest candidate value. Overall, wrong predictions are typically characterized by inconsistent candidate preferences and a lack of robust discriminative advantage for the correct candidate.


Overall, through a comprehensive analysis of attention allocation, hidden-state similarity, and output logits, we find that the success or failure of latest-state retrieval is not determined by a single layer or a single local signal. On the contrary, these cues are more likely to exist in a cross-layer, distributed, and synergistic manner, and they interact and couple across different representation layers to manifest as an integrated, holistic feature.

\begin{table*}[t]
\centering
\setlength{\tabcolsep}{4.5pt}
\renewcommand{\arraystretch}{1.15}
\resizebox{\textwidth}{!}{%
\begin{tabular}{ll c c c c c c c c}
\toprule
\multirow{2}{*}{\textbf{Model}} & \multirow{2}{*}{\textbf{Query}} & \multirow{2}{*}{\textbf{WO Intervention}}
& \multicolumn{3}{c}{\textbf{General Prompting}}
& \multicolumn{2}{c}{\textbf{General Mnemonic}}
& \multicolumn{2}{c}{\textbf{Memory Updating}} \\
\cmidrule(lr){4-6}\cmidrule(lr){7-8}\cmidrule(lr){9-10}
& & & CoT & 2-Shot & Index & Rehearsal & Paraphrase & Integration & Forgetting \\
\midrule
\multirow{3}{*}{\textbf{LLaMA3.1-8B}}
& Earliest & \mstd{98.33}{1.70}  & \mstd{95.83}{0.62} & \mstd{\textcolor[HTML]{104680}{\textbf{99.67}}}{0.24} & \mstd{96.67}{1.43} & \mstd{96.33}{1.03} & \mstd{\textcolor[HTML]{8A2230}{\textbf{99.33}}}{0.47} & \mstd{98.00}{0.82}   & \mstd{98.00}{0.82} \\
& Latest   & \mstd{4.50}{0.41}   & \mstd{4.33}{0.94}  & \mstd{\textcolor[HTML]{104680}{\textbf{8.67}}}{0.24}  & \mstd{2.83}{0.47}  & \mstd{4.17}{0.62}  & \mstd{5.00}{0.0}        & \mstd{4.83}{0.85} & \mstd{\textcolor[HTML]{8A2230}{\textbf{5.83}}}{0.85} \\
& ELAG      & 93.83              & 91.50               & 91.00                 & 93.84              & 92.16              & 94.33              & 93.17              & 92.17 \\
\midrule

\multirow{3}{*}{\textbf{LLaMA3.1-70B}}
& Earliest & \mstd{100.00}{0.00}                & \mstd{100.00}{0.00}      & \mstd{99.83}{0.24} & \mstd{\textcolor[HTML]{104680}{\textbf{100.00}}}{0.00}      & \mstd{99.83}{0.24} & \mstd{100.00}{0.00}      & \mstd{100.00}{0.00}      & \mstd{\textcolor[HTML]{8A2230}{\textbf{100.00}}}{0.00} \\
& Latest   & \mstd{72.17}{1.7}  & \mstd{71.50}{2.16}  & \mstd{\textcolor[HTML]{104680}{\textbf{76.50}}}{0.41}  & \mstd{38.33}{3.17} & \mstd{72.50}{0.71}  & \mstd{\textcolor[HTML]{8A2230}{\textbf{72.67}}}{2.49} & \mstd{68.67}{0.47} & \mstd{\textcolor[HTML]{8A2230}{\textbf{72.67}}}{2.09} \\
& ELAG      & 27.83              & 28.50               & 23.33              & 61.67              & 27.33              & 27.33              & 31.33              & 27.33 \\
\midrule

\multirow{3}{*}{\textbf{Qwen2.5-7B}}
& Earliest & \mstd{99.33}{0.62} & \mstd{\textcolor[HTML]{104680}{\textbf{100.00}}}{0.00}      & \mstd{97.50}{0.82}  & \mstd{99.17}{0.24} & \mstd{99.67}{0.24} & \mstd{100.00}{0.00}      & \mstd{\textcolor[HTML]{8A2230}{\textbf{100.00}}}{0.00}      & \mstd{98.00}{0.82} \\
& Latest   & \mstd{18.67}{0.47} & \mstd{19.5}{0.41}  & \mstd{\textcolor[HTML]{104680}{\textbf{22.67}}}{2.39} & \mstd{17.83}{2.25} & \mstd{19.83}{1.43} & \mstd{20.5}{2.04}  & \mstd{\textcolor[HTML]{8A2230}{\textbf{25.83}}}{2.46} & \mstd{21.5}{1.63} \\
& ELAG      & 80.66              & 80.50               & 74.83              & 81.34              & 79.84              & 79.50               & 74.17              & 76.50 \\
\midrule

\multirow{3}{*}{\textbf{Qwen2.5-14B}}
& Earliest & \mstd{100.00}{0.00}                & \mstd{100.00}{0.00}                & \mstd{99.83}{0.24} & \mstd{\textcolor[HTML]{104680}{\textbf{100.00}}}{0.00}      & \mstd{100.00}{0.00}      & \mstd{99.83}{0.24} & \mstd{\textcolor[HTML]{8A2230}{\textbf{100.00}}}{0.00}      & \mstd{99.83}{0.24} \\
& Latest   & \mstd{27.00}{0.71}    & \mstd{30.17}{1.25} & \mstd{30.83}{1.03} & \mstd{\textcolor[HTML]{104680}{\textbf{47.00}}}{4.14}    & \mstd{32.67}{1.03} & \mstd{35.00}{3.19}    & \mstd{33.67}{3.42} & \mstd{\textcolor[HTML]{8A2230}{\textbf{35.50}}}{3.34} \\
& ELAG      & 73.00                 & 69.83              & 69.00                 & 53.00                 & 67.33              & 64.83              & 66.33              & 64.33 \\
\midrule

\multirow{3}{*}{\textbf{Qwen2.5-32B}}
& Earliest & \mstd{100.00}{0.00}                & \mstd{100.00}{0.00}                & \mstd{100.00}{0.00}                & \mstd{100.00}{0.00}                & \mstd{100.00}{0.00}                & \mstd{100.00}{0.00}                & \mstd{100.00}{0.00}                & \mstd{\textcolor[HTML]{8A2230}{\textbf{100.00}}}{0.00} \\
& Latest   & \mstd{57.33}{3.70} & \mstd{56.50}{1.87}  & \mstd{51.67}{1.03} & \mstd{56.17}{1.31} & \mstd{57.83}{1.31} & \mstd{55.00}{2.48}    & \mstd{57.17}{0.62} & \mstd{\textcolor[HTML]{8A2230}{\textbf{58.33}}}{0.94} \\
& ELAG      & 42.67              & 43.50               & 48.33              & 43.83              & 42.17              & 45.00                 & 42.83              & 41.67 \\
\midrule

\multirow{3}{*}{\textbf{Qwen2.5-72B}}
& Earliest & \mstd{100.00}{0.00}                & \mstd{100.00}{0.00}                & \mstd{\textcolor[HTML]{104680}{\textbf{100.00}}}{0.00}                & \mstd{100.00}{0.00}                & \mstd{100.00}{0.00}                & \mstd{100.00}{0.00}                & \mstd{100.00}{0.00}                & \mstd{\textcolor[HTML]{8A2230}{\textbf{100.00}}}{0.00} \\
& Latest   & \mstd{55.33}{2.9}  & \mstd{53}{2.94}    & \mstd{\textcolor[HTML]{104680}{\textbf{66.67}}}{1.43} & \mstd{27.00}{2.16}    & \mstd{54.67}{3.7}  & \mstd{55.17}{3.57} & \mstd{52.00}{1.63}    & \mstd{\textcolor[HTML]{8A2230}{\textbf{56.00}}}{2.27} \\
& ELAG      & 45.67              & 47.00                 & 33.33              & 73.00                 & 45.33              & 44.83              & 48.00                 & 44.00 \\
\midrule

\multirow{3}{*}{\textbf{Qwen3-235B}}
& Earliest & \mstd{100.00}{0.00}                 & \mstd{99.83}{0.24} & \mstd{\textcolor[HTML]{104680}{\textbf{100.00}}}{0.00}                & \mstd{100.00}{0.00}                & \mstd{100.00}{0.00}                & \mstd{100.00}{0.00}                & \mstd{\textcolor[HTML]{8A2230}{\textbf{100.00}}}{0.00}                & \mstd{99.83}{0.24} \\
& Latest   & \mstd{80.67}{0.94} & \mstd{79.83}{1.93} & \mstd{\textcolor[HTML]{104680}{\textbf{84.00}}}{2.16}    & \mstd{66.00}{3.49}    & \mstd{78.17}{2.78} & \mstd{78.33}{2.25} & \mstd{\textcolor[HTML]{8A2230}{\textbf{86.17}}}{1.03} & \mstd{81.33}{1.84} \\
& ELAG      & 19.33              & 20.00                 & 16.00                 & 34.00                 & 21.83              & 21.67              & 13.83              & 18.50 \\
\bottomrule
\end{tabular}%
}
\caption{Performance (\%) on the synthetic dataset. Results are reported as mean$_{\mathrm{std}}$, and ELAG is the mean difference between earliest and latest accuracy. Blue highlights the best result among general prompting methods, while red highlights the best result among cognitively inspired methods (General Mnemonic + Memory Updating). The ELAG metric highlights a systematic bias: while earliest-state retrieval is near ceiling across settings, latest-state retrieval remains the dominant bottleneck even under general prompting and heuristic interventions.}
\label{tab:rw_interventions_mean_std}
\end{table*}

\section{Interventions}

\subsection{The Prompt Design of Different Intervention Method}

\label{Appendix:D.1}

\noindent \textbf{General mnemonic strategies}.

\noindent \textbf{1. Rote rehearsal prompting}: The rote rehearsal intervention strengthens retention by repeatedly restating newly acquired information \cite{hartshorne2019effect,plaska2021does}. Glenberg and Adams \shortcite{glenberg1978type} regarded rehearsal as a fundamental mnemonic strategy: repetition stabilizes information representations during short-term maintenance and influences subsequent memory performance, making it particularly suitable for tasks that require sustained information retention. Its core mechanism lies in the repeated activation of the same content, thereby improving the accuracy of subsequent retrieval.

\begin{tcolorbox}[
    colback=gray!2,
    arc=5pt,
    left=10pt,
    right=10pt,
    top=6pt,
    bottom=6pt,
    title= Rote rehearsal prompting,
    fonttitle=\large,
    center title,
    breakable,
]
You are given a long updated list of \dots .\\[6pt]

\textbf{Rehearse each new cue:value pair three times when you read it. Do this internally and do not output the rehearsals into text.}

\textbf{CUE (JSON array):} \texttt{["President of Italy"]}\\[8pt]

\textbf{INPUT FORMAT}\\[-3pt]

\dots
\dots
\end{tcolorbox}

\noindent \textbf{2. Semantic elaboration prompting}: The levels-of-processing theory \cite{craik1972levels} argues that memory performance depends not only on the amount of rehearsal but also on the depth of processing during encoding \cite{bartsch2021effects,nieznanski2020levels}. Compared with shallow perceptual processing, deeper semantic elaboration, such as grasping meaning, forming semantic associations, and generating explanations or contexts, typically yields stronger retention. Its core mechanism is to enrich interpretive representations and link new information to existing knowledge structures, making it easier to recover during subsequent retrieval.

\begin{tcolorbox}[
    colback=gray!2,
    arc=5pt,
    left=10pt,
    right=10pt,
    top=6pt,
    bottom=6pt,
    title= Semantic elaboration prompting,
    fonttitle=\large,
    center title,
    breakable,
]
You are given a long updated list of \dots .\\[6pt]

\textbf{Create a meaningful association between the next cue:value pair and the previous cue:value pair based on semantic meaning. Create these associations internally and do not output it in your response.}

\textbf{CUE (JSON array):} \texttt{["President of Italy"]}\\[8pt]

\textbf{INPUT FORMAT}\\[-3pt]

\dots
\dots
\end{tcolorbox}

\noindent \textbf{Memory-updating strategies.}

\noindent \textbf{1. Memory integration prompting.} Memory integration emphasizes that when new information coexists or overlaps with existing memories, co-activating both old and new information during the learning and encoding phases and integrating them into a more coherent representation helps reduce interference from outdated information in subsequent retrieval \cite{morton2023memory,molitor2021memory}. Schlichting and Preston \shortcite{schlichting2015memory} presented relevant evidence indicating that the co-activation of prior knowledge and new inputs can facilitate the integration process, thereby influencing subsequent behavioral and retrieval performance. Based on this insight, our prompt explicitly encourages the model to organize sequentially updated cue:value pairs into an update chain \cite{chanales2019interference}, treating the DKI sequence as an interconnected trajectory rather than a set of independent pairs. By encoding the relational structure between updates (i.e., how each new value revises the previous one), the model is guided to retrieve the latest state as the terminal link of the chain while suppressing earlier values that have been superseded by subsequent updates.

\begin{tcolorbox}[
    colback=gray!2,
    arc=5pt,
    left=10pt,
    right=10pt,
    top=6pt,
    bottom=6pt,
    title= Memory integration prompting,
    fonttitle=\large,
    center title,
    breakable,
]
You are given a long updated list of \dots .\\[6pt]

\textbf{Internal organization rule:}\\[-3pt]
\begin{itemize}
    \item As you read from top to bottom, maintain for each cue in CUES an evolving chain:\\
    \hspace*{1em}\texttt{CUE: v(1) $\rightarrow$ v(2) $\rightarrow$ $\cdot$ $\rightarrow$ v(T)},\\
    where \texttt{v(1)} is the value from the first occurrence and \texttt{v(T)} is the value from the last occurrence.
    \item Do not treat records as independent pairs. Treat each later occurrence of the same cue as an update to that cue, so that the earliest value is \texttt{v(1)} and the latest value is \texttt{v(T)}.
\end{itemize}

\textbf{CUE (JSON array):} \texttt{["President of Italy"]}\\[8pt]

\textbf{INPUT FORMAT}\\[-3pt]

\dots
\dots
\end{tcolorbox}

\noindent \textbf{2. Directed forgetting prompting.} Directed forgetting \cite{macleod1999item} is a strategy that reduces access to outdated information through explicit instructions. After new information replaces old information, earlier representations are deliberately suppressed so that they are less likely to intrude during later retrieval. The core principle is to mark old information as obsolete or no longer applicable, thereby lowering the probability that it will be erroneously retrieved \cite{fellner2020tracking,abel2021list}.

\begin{tcolorbox}[
    colback=gray!2,
    arc=5pt,
    left=10pt,
    right=10pt,
    top=6pt,
    bottom=6pt,
    title= Directed forgetting prompting,
    fonttitle=\large,
    center title,
    breakable,
]
You are given a long updated list of \dots .\\[6pt]

\textbf{When reading the list of cue-value records, overwrite every previous cue-value pair with the current cue-value pair.}

\textbf{CUE (JSON array):} \texttt{["President of Italy"]}\\[8pt]

\textbf{INPUT FORMAT}\\[-3pt]

\dots
\dots
\end{tcolorbox}

\noindent \textbf{General Prompting Strategies}

In addition to our cognitively inspired interventions, we include several widely used general prompting strategies as baselines to assess whether general prompt engineering alone can alleviate retrieval bias under multi-update settings. Specifically, we consider Chain-of-Thought (CoT), 2-shot prompting, and an Index-style structured prompt.

\noindent\textbf{1. CoT Prompting.} For the \textbf{CoT prompting} setting \cite{wei2022chain}, we encourage hidden step-by-step reasoning via the following chain-of-thought instruction:

\begin{tcolorbox}[
    colback=gray!2,
    arc=5pt,
    left=10pt,
    right=10pt,
    top=6pt,
    bottom=6pt,
    title= CoT Prompting,
    fonttitle=\large,
    center title,
    breakable,
]
You are given \dots.\\[6pt]

\textbf{Chain-of-Thought (CoT) instructions:}\\[-3pt]
\begin{itemize}
    \item Think step by step to solve the task, but keep all reasoning hidden.
    \item Do not output any explanations or reasoning; output only the final JSON object.
\end{itemize}

\textbf{CUE (JSON array):} \texttt{["President of Italy"]}\\[8pt]

\textbf{INPUT FORMAT}\\[-3pt]

\dots

\dots
\end{tcolorbox}

\noindent\textbf{2. Few-shot Prompting.} For the \textbf{few-shot prompting} setting \cite{brown2020language}, we provide solved examples and then ask the model to follow the same pattern using the following prompt:

\begin{tcolorbox}[
    colback=gray!2,
    arc=5pt,
    left=10pt,
    right=10pt,
    top=6pt,
    bottom=6pt,
    title= Few-shot Prompting,
    fonttitle=\large,
    center title,
    breakable,
]

You are given \dots.\\[6pt]

Below are two examples. Follow the same pattern:\\[6pt]

\textbf{EXAMPLE 1}\\
\texttt{START:}\\
\texttt{edgewise:artistic}\\
\texttt{edgewise:tributes}\\
\texttt{edgewise:overplay}\\
\texttt{edgewise:cowardly}\\
\texttt{edgewise:applause}\\
\texttt{edgewise:slavered}\\
\texttt{edgewise:coincide}\\
\texttt{edgewise:teletype}\\
\texttt{edgewise:sunburnt}\\
\texttt{END}\\[4pt]

Correct output:\\
\texttt{\{}\\
\texttt{~~"cue":"edgewise",}\\
\texttt{~~"latest":"sunburnt",}\\
\texttt{~~"earliest":"artistic"}\\
\texttt{\}}\\[8pt]

\textbf{EXAMPLE 2}\\
\texttt{START:}\\
\texttt{tributes:coherent}\\
\texttt{tributes:allergen}\\
\texttt{tributes:shivered}\\
\texttt{tributes:cowardly}\\
\texttt{tributes:arranged}\\
\texttt{tributes:emeritus}\\
\texttt{tributes:teletype}\\
\texttt{tributes:antennae}\\
\texttt{END}\\[4pt]

Correct output:\\
\texttt{\{}\\
\texttt{~~"cue":"tributes",}\\
\texttt{~~"latest":"antennae",}\\
\texttt{~~"earliest":"coherent"}\\
\texttt{\}}\\[8pt]

Now solve the following instance in exactly the same way.\\

\textbf{CUE (JSON array):} \texttt{["President of Italy"]}\\[8pt]

\textbf{INPUT FORMAT}\\[-3pt]

\dots
\dots
\end{tcolorbox}

\noindent\textbf{3. Index-based Prompting.} For the \textbf{index-based prompting} setting, we explicitly number each update step, as illustrated in the following prompt:

\begin{tcolorbox}[
    colback=gray!2,
    arc=5pt,
    left=10pt,
    right=10pt,
    top=6pt,
    bottom=6pt,
    title= Index-based Prompting,
    fonttitle=\large,
    center title,
    breakable,
]

You are given \dots.\\[6pt]

\textbf{CUE (JSON array):} \texttt{["President of Italy"]}\\[8pt]

\textbf{INPUT FORMAT}\\[-3pt]

- Each record is one line in the form: \texttt{cue:value}

- Boundaries: lines strictly between the literal markers \texttt{START:} and \texttt{END}

- Index is a monotonically increasing integer (0, 1, 2, ...). It only indicates the order;

\textbf{Record List}\\[-3pt]

\texttt{START:}\\
\texttt{1.~President of Italy:Alcide De Gasperi}\\
\texttt{2.~President of Italy:Enrico de Nicola}\\
\texttt{3.~President of Italy:Luigi Einaudi}\\
\texttt{4.~President of Italy:Giovanni Gronchi}\\
\texttt{5.~President of Italy:Antonio Segni}\\
\texttt{6.~President of Italy:Giuseppe Saragat}\\
\texttt{7.~President of Italy:Giovanni Leone}\\
\texttt{8.~President of Italy:Sandro Pertini}\\
\texttt{9.~President of Italy:Francesco Cossiga}\\
\texttt{10.~President of Italy:Oscar Luigi Scalfaro}\\
\texttt{11.~President of Italy:Carlo Azeglio Ciampi}\\
\texttt{12.~President of Italy:Giorgio Napolitano}\\
\texttt{13.~President of Italy:Sergio Mattarella}\\
\texttt{END}\\

\dots
\dots
\end{tcolorbox}

\subsection{Heuristic Interventions}

\label{Appendix:D.2}

Given that local internal signals do not reveal an stable and effective angle for internal intervention, we introduce cognitively motivated heuristic interventions that use prompting to strengthen the model's internal encoding of the latest update or to reduce interference from competition among multiple candidate values. To evaluate the effectiveness of different prompt strategies and cognitive heuristic-based interventions in mitigating retrieval bias, we conduct experiments on the synthetic DKI dataset. 
Table~\ref{tab:rw_interventions_mean_std} presents the experimental results when the number of updates is set to $T=32$. At this update length, the retrieval bias phenomenon can be stably observed. We perform experiments and comparative analyses on two mainstream open-source model families, namely LLaMA 3.1 and Qwen 2.5/3.

The overall results still exhibit retrieval bias: regardless of model scale or intervention prompt variants, earliest-state retrieval performance is nearly at ceiling level (typically around 98-100\%), indicating that earliest historical state retrieval remains highly accessible and is largely unaffected by prompt strategies. In contrast, latest-state retrieval performance fluctuates significantly and consistently constitutes the primary bottleneck, thereby resulting in a substantial earliest-latest accuracy gap across most experimental settings.

Among general prompt strategies, 2-shot prompting stands out as one of the most stable and consistently effective methods for improving latest-state retrieval accuracy (e.g., LLaMA3.1-8B: 4.50\%$\rightarrow$8.67\%; LLaMA3.1-70B: 72.17\%$\rightarrow$76.50\%; Qwen2.5-72B: 55.33\%$\rightarrow$66.67\%). In contrast, the Index strategy exhibits a rather counterintuitive fragility and may even significantly degrade latest-state retrieval performance (e.g., dropping to 38.33\% for LLaMA3.1-70B; falling to 27.00\% for Qwen2.5-72B). This could be attributed to the fact that explicit sequential numbering introduces additional interferential tokens, which alter the original cue:value pattern, thereby diverting the model's attention and focus away from candidate values and ultimately resulting in unstable latest-state retrieval.

Among cognitive heuristic interventions, memory integration and directed forgetting tend to yield more stable performance gains.
Memory integration achieves the optimal performance on certain models (e.g., Qwen2.5-7B: 18.67\%$\rightarrow$25.83\%; Qwen3-235B: 80.67\%$\rightarrow$86.17\%), while directed forgetting also demonstrates competitiveness on some models (e.g., Qwen2.5-32B: 57.33\%$\rightarrow$58.33\%; Qwen2.5-72B: 55.33\%$\rightarrow$56.00\%). In addition, the improvements in latest-state retrieval accuracy outperform those of general mnemonic strategies.


Although various strategies have improved latest-state retrieval to some extent, the earliest-latest accuracy gap  remains remarkably pronounced for small models (e.g., still approximately 91\% for LLaMA3.1-8B). Overall, general prompt strategies and cognitive heuristic interventions can mitigate the bottleneck in latest-state retrieval to a certain degree, yet they are insufficient to fully eliminate the underlying retrieval bias. Future research should further advance efforts from both the model side and the task side in a synergistic manner, to enhance the model's ability to track the latest evidence and thereby systematically narrow the earliest-latest accuracy gap.

\end{document}